\documentclass[runningheads]{llncs}
\usepackage{graphicx}
\usepackage{comment}
\usepackage{amsmath,amssymb} % define this before the line numbering.
\usepackage{color}
 \usepackage{mathrsfs}
 \usepackage{subfigure}
\begin{document}
% \linenumbers
\pagestyle{headings}
\mainmatter
\def\ECCVSubNumber{184}  % Insert your submission number here

\title{Image Stitching and Rectification \\for Hand-Held Cameras} % Replace with your title

\begin{comment}
\titlerunning{ECCV-20 submission ID \ECCVSubNumber} 
\authorrunning{ECCV-20 submission ID \ECCVSubNumber} 
\author{Anonymous ECCV submission}
\institute{Paper ID \ECCVSubNumber}
\end{comment}
%******************

%\begin{comment}
\titlerunning{Image Stitching and Rectification for Hand-Held Cameras}
\author{Bingbing Zhuang\orcidID{0000-0002-2317-3882} \and Quoc-Huy Tran\orcidID{0000-0003-1396-6544}}
\authorrunning{B. Zhuang and  Q.-H. Tran}
\institute{NEC Labs America
}
%\end{comment}
%******************
\maketitle

\begin{abstract}
In this paper, we derive a new differential homography that can account for the scanline-varying camera poses in Rolling Shutter (RS) cameras, and demonstrate its application to carry out RS-aware image stitching and rectification at one stroke. Despite the high complexity of RS geometry, we focus in this paper on a special yet common input --- two consecutive frames from a video stream, wherein the inter-frame motion is restricted from being arbitrarily large. This allows us to adopt simpler differential motion model, leading to a straightforward and practical minimal solver. To deal with non-planar scene and camera parallax in stitching, we further propose an RS-aware spatially-varying homogarphy field in the principle of As-Projective-As-Possible (APAP). We show superior performance over state-of-the-art methods both in RS image stitching and rectification, especially for images captured by hand-held shaking cameras. %, which pose great challenges to existing methods.

\end{abstract}

\section{Introduction}
\label{sec:introduction}

\begin{figure}[t]
	\centering
		\includegraphics[width=1.0\linewidth, trim = 0mm 38mm 70mm 0mm, clip]{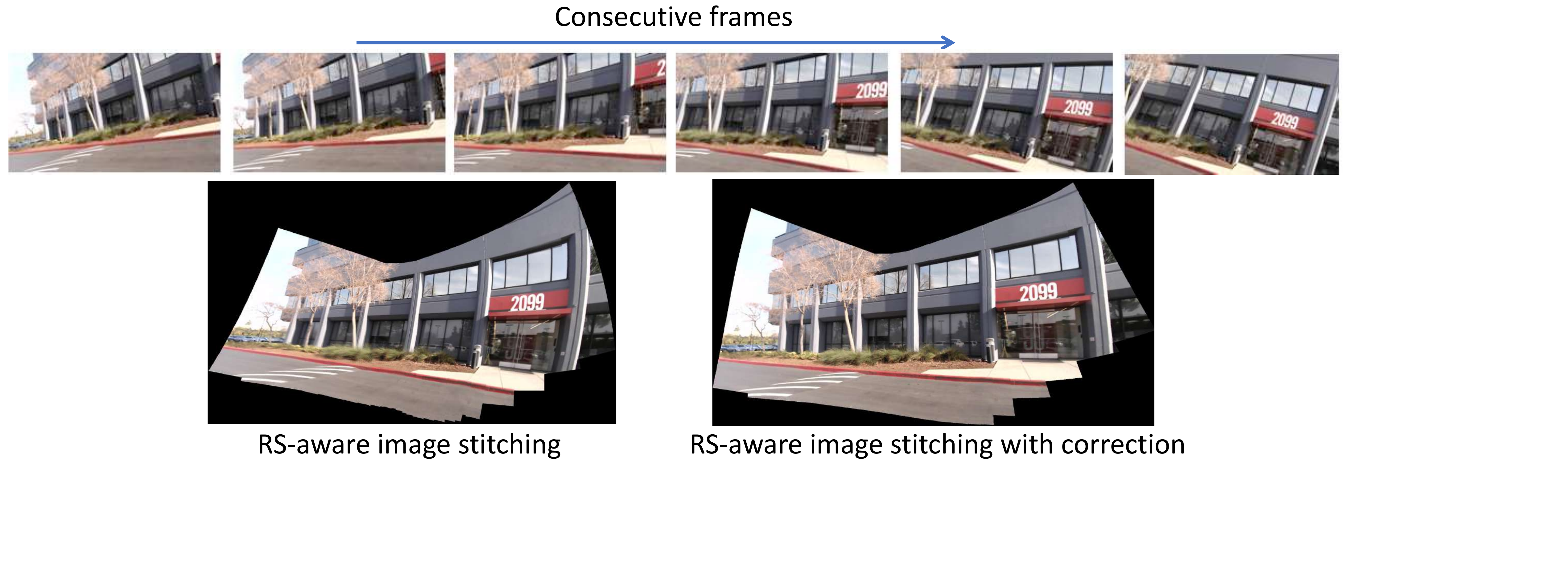}
	\caption{Example results of RS-aware image stitching and rectification.}
	\label{fig:teaser}
\end{figure}

Rolling Shutter (RS) cameras adopt CMOS sensors due to their low cost and simplicity in manufacturing. This stands in contrast to Global Shutter (GS) CCD cameras that require specialized and highly dedicated fabrication. Such discrepancy endows RS cameras great advantage for ubiquitous employment in consumer products, e.g., smartphone cameras~\cite{muratov20163dcapture} or dashboard cameras~\cite{haresh2020towards}. However, the expediency in fabrication also causes a serious defect in image capture --- instead of capturing different scanlines all at once as in GS cameras, RS cameras expose each scanline one by one sequentially from top to bottom. While static RS camera capturing a static scene is fine, the RS effect comes to haunt us as soon as images are taken during motion, i.e., images could be severely distorted due to scanline-varying camera poses (see Fig.~\ref{fig:teaser}).

RS distortion has been rearing its ugly head in various computer vision tasks. %(e.g.~\cite{saurer2013rolling,hedborg2012rolling,dai2016rolling,zhuang2017rolling,albl2015r6p,lao2018robust,purkait2017roll%ing,laopami20,rengarajan2016bows}). 
There is constant pressure to either remove the RS distortion in the front-end image capture~\cite{lao2018robust,purkait2017rolling,rengarajan2016bows,vasu2018occlusion}, or design task-dependent RS-aware algorithms in the back end~\cite{saurer2013rolling,hedborg2012rolling,dai2016rolling,albl2015r6p,mohan2017going,rengarajan2016image,punnappurath2015rolling}. While various algorithms have been developed for each of them in isolation, algorithms achieving both in a holistic way are few~\cite{ringaby2012efficient,zhuang2017rolling,laopami20,vasu2017camera}. In this paper, we make contributions towards further advancement in this line. Specifically, we propose a novel differential homography and demonstrate its application to carry out RS image stitching and rectification at one stroke.  

RS effect complicates the two-view geometry significantly compared to its GS counterpart, primarily because 12 additional unknown parameters are required to model the intra-frame velocity of the two cameras. Thus, despite the recent effort of Lao et al.~\cite{laopami20} in solving a generic RS homography for discrete motion, the complexity of RS geometry significantly increases the number of required correspondences (36 points for full model and 13.5 points after a series of approximations).
Inspired by prior work~\cite{zhuang2017rolling} that demonstrates dramatic simplification in differential RS relative pose estimation compared to its discrete counterpart~\cite{dai2016rolling},   
%In Dai et al.~\cite{dai2016rolling}, it has been shown that the generic RS two-view geometry becomes much more complicated than its GS counterpart due to the extra 12 unknown parameters to model the intra-frame velocity.
%However, the later work~\cite{zhuang2017rolling} demonstrates dramatically reduced complexity under small motion assumption by using instantaneous motion model. Following this spirit, 
we focus in this paper on the special yet common case where the inputs are two consecutive frames from a video. In this case, the inter-frame motion is restricted from being arbitrarily large, allowing us to adopt the simpler differential homography model~\cite{ma2012invitation}. Furthermore, the intra-frame motion could be directly parameterized by the inter-frame motion via interpolation, thereby reducing the total number of unknown parameters to solve. In particular, we derive an RS-aware differential homography under constant acceleration motion assumption, together with a straightforward solver requiring only 5 pairs of correspondences, and demonstrate its application to simultaneous RS image stitching and rectification. 
Since a single homography warping is only exact under pure rotational camera motion or for 3D planar scene, it often causes misalignment when such condition is not strictly met in practice.
%There are constant efforts towards handling camera parallax or non-planar scene in image stitching research. 
%We show that the simplicity of our differential homography permits it to be readily integrated into more advanced image stitching methods. 
To address such model inadequacy, we extend the single RS homography model to a spatially-varying RS homography field following the As-Projective-As-Possible (APAP) principle~\cite{zaragoza2013projective}, thereby lending itself to handling complex scenes. We demonstrate example results in Fig.~\ref{fig:teaser}, where multiple images are stitched and rectified by concatenating pairwise warping from our method.

%Close to our work is the generic RS homography included in Lao's thesis~\cite{laopami20} for discrete motion. However, our differential formulation for continuous motion leads to a much simpler and more practical minimal solver, yet obtain strong performance in practice. 
%Our image rectification bears resemblance to the SfM-based method~\cite{zhuang2017rolling}, in the sense that both apply differential motion model. 
We would also like to emphasize our advantage over the differential Structure-from-Motion (SfM)-based rectification method~\cite{zhuang2017rolling}. Note that \cite{zhuang2017rolling} computes the rectification for each pixel separately via pixel-wise depth estimation from optical flow and camera pose. As such, potential gross errors in optical flow estimates could lead to severe artifacts in the texture-less or non-overlapping regions. In contrast, the more parsimonious homography model offers a natural defense against wrong correspondences. Despite its lack of full 3D reconstruction, we observe good empirical performance  in terms of visual appearance.

In summary, our contributions include:
\begin{itemize}
\item We derive a novel differential homography model together with a minimal solver to account for the scanline-varying camera poses of RS cameras.
\item We propose an RS-aware spatially-varying homography field for improving RS image stitching.  
\item Our proposed framework outperforms state-of-the-art methods both in RS image rectification and stitching.
\end{itemize}
\section{Related Work}
\label{sec:relatedwork}

\noindent \textbf{RS Geometry.}
Since the pioneering work of Meingast et al.~\cite{meingast2005geometric}, considerable efforts have been invested in studying the geometry of RS cameras. These include relative pose estimation~\cite{dai2016rolling,zhuang2017rolling,purkait2018minimal}, absolute pose estimation~\cite{magerand2012global,albl2015r6p,saurer2015minimal,lao2018rolling,Albl_2016_CVPR,kukelova2018linear}, bundle adjustment~\cite{hedborg2012rolling,klingner2013street}, SfM/Reconstruction~\cite{saurer2013rolling,im2015high,saurer2016sparse,Schubert_2018_ECCV,schubert2018direct}, degeneracies~\cite{albl2016degeneracies,ito2017self,zhuang2019learning}, discrete homography~\cite{laopami20}, and others~\cite{bapat2018rolling,oth2013rolling}. In this work, we introduce RS-aware differential homography, which is of only slighly higher complexity than its GS counterpart. %which is useful for tasks such as RS image stitching and rectification.

\noindent \textbf{RS Image Rectification.}
%First, there is a line of research that targets at rectification from a single RS image. 
Removing RS artifacts using a \textit{single} input image is inherently an ill-posed problem. Works in this line ~\cite{rengarajan2016bows,purkait2017rolling,lao2018robust} often assume simplified camera motions and scene structures, and require line/curve detection in the image, if available at all. Recent methods~\cite{rengarajan2017unrolling,zhuang2019learning} have started exploring deep learning for this task. However, their generalization ability to different scenes remains an open problem.
In contrast, \textit{multi-view} approaches, be it geometric-based or learning-based~\cite{Liu_2020_CVPR}, are more geometrically grounded. 
 %Therefore, classical single-view methods often assume simplified camera motions and scene structures, e.g., pure rotation~\cite{rengarajan2016bows,purkait2017rolling,lao2018robust} and Manhattan world~\cite{purkait2017rolling}. Moreover, they require a sufficient number of lines/curves correctly detected in the input image to work well. Recently, deep learning-based single-view approaches~\cite{rengarajan2017unrolling,zhuang2019learning} have emerged as promising alternatives, however, their generalization ability across different scenes such as indoor and outdoor scenes has not been validated thoroughly.
%To avoid the ill-posedness of single-view methods, multiple-view approaches rely on two or more input images. 
In particular, %given sparse correspondences computed between the input images, 
Ringaby and Forssen~\cite{ringaby2012efficient} estimate and smooth a sequence of camera rotations for eliminating RS distortions, while Grundmann et al.~\cite{grundmann2012calibration} and Vasu et al.~\cite{vasu2018occlusion} use a mixture of homographies to model and remove RS effects. 
%These iterative optimization-based methods either assume no outlier correspondences~\cite{ringaby2012efficient} or alleviate their effects by adding a regularizer~\cite{grundmann2012calibration} or robust loss~\cite{vasu2018occlusion}. 
Such methods often rely on nontrivial iterative optimization leveraging a large set of correspondences. 
Recently, Zhuang et al.~\cite{zhuang2017rolling} present the first attempt to derive minimal solver for RS rectification. It takes a minimal set of points as input and lends itself well to RANSAC, leading to a more principled way for robust estimation. In the same spirit, we derive RS-aware differential homography and show important advantages. Note that our minimal solver is orthogonal to the optimization-based methods, e.g.~\cite{ringaby2012efficient,vasu2018occlusion}, and can serve as their initialization. Very recently, Albl et al.~\cite{albl2020two} present an interesting way for RS undistortion from two cameras, yet require specific camera mounting. 

%In this work, we introduce a more principled way to handle outlier correspondences, i.e., minimal solver incorporated within RANSAC for robust estimation. 

%Recently, Zhuang et al.~\cite{zhuang2017rolling} extract dense correspondences between the images and explore RS geometry to estimate both camera motion and scene structure, which are then used for removing RS distortions. In contrast to the full 3D reconstruction-based approach of~\cite{zhuang2017rolling}, our homography-based method does not require dense correspondences, and is able to generalize well to texture-less or non-overlapping regions (see Sec.~\ref{sec:rsrectification} for a more detailed discussion).

\noindent \textbf{GS Image Stitching.}
Image stitching~\cite{szeliski2007image} has achieved significant progress over the past few decades. % with the release of commercialized products such as Autostitch and Photosynth. 
Theoretically, a single homography is sufficient to align two input images of a common scene if the images are captured with no parallax or the scene is planar~\cite{hartley2003multiple}. In practice, this condition is often violated, causing misalignments or ghosting artifacts in the stitched images. To overcome this issue, several approaches have been proposed such as spatially-varying warps~\cite{liu2009content,lin2011smoothly,zaragoza2013projective,lee2020warping,liao2019single}, shape-preserving warps~\cite{chang2014shape,chen2016natural,lin2015adaptive}, and seam-driven methods~\cite{zhang2014parallax,lin2016seagull,herrmann2018robust,herrmann2018object}. All of the above approaches assume a GS camera model and hence they cannot handle RS images, i.e., the stitched images may contain RS distortion-induced misalignment. While Lao et al.~\cite{laopami20} demonstrate the possibility of stitching in spite of RS distortion, we present a more concise and straightforward method that works robustly with hand-held cameras. 
%In contrast, we model RS distortions explicitly and provide a more geometrically faithful way for RS image stitching and rectification.
\section{Homography Preliminary}
\label{sec:background}

%\subsection{GS Discrete Homography}
\noindent \textbf{GS Discrete Homography.} Let us assume that two calibrated cameras are observing a 3D plane parameterized as $(\boldsymbol{n},d)$, with $\boldsymbol{n}$ denoting the plane normal and $d$ the camera-to-plane distance. Denoting the relative camera rotation and translation as $\boldsymbol{R} \in SO(3)$ and $\boldsymbol{t}\in \mathbb{R}^3$, a pair of 2D correspondences $\boldsymbol{x}_1$ and $\boldsymbol{x}_2$ (in normalized plane) can be related by $\hat{\boldsymbol{x}}_2 \propto \boldsymbol{H}\hat{\boldsymbol{x}}_1$,
where $\boldsymbol{H} = \boldsymbol{R}+{\boldsymbol{tn}^\top}/{d}$ is defined as the \emph{discrete} homography~\cite{hartley2003multiple} and $\hat{\boldsymbol{x}}=[\boldsymbol{x}^\top,1]^\top$. $\propto$ indicates equality up to a scale. Note that $\boldsymbol{H}$ in the above format  subsumes the pure rotation-induced homography as a special case by letting $d\rightarrow\infty$. 
Each pair of correspondence $\{\boldsymbol{x}_1^i,\boldsymbol{x}_2^i\}$ gives two constraints $\boldsymbol{a}_i\boldsymbol{h}=\boldsymbol{0}$, where $\boldsymbol{h} \in \mathbb{R}^{9}$ is the vectorized form of $\boldsymbol{H}$ and the coefficients $\boldsymbol{a}_i\in \mathbb{R}^{2\times 9}$ can be computed from $\{\boldsymbol{x}_1^i,\boldsymbol{x}_2^i\}$. In \emph{GS discrete 4-point solver}, with the minimal of 4 points, one can solve $\boldsymbol{h}$ via:

%With a minimal of 4 pairs of correspondences $\{\boldsymbol{x}_1^i,\boldsymbol{x}_2^i\}$, one can estimate $\boldsymbol{H}$ by solving: 
\begin{equation}
%\boldsymbol{h}^* = \arg\min_{\boldsymbol{h}} \sum_i \|\boldsymbol{a}_i\boldsymbol{h}\|^2 = %\arg\min_{\boldsymbol{h}} \|\boldsymbol{Ah}\|^2,~~~s.t.~ \|\boldsymbol{h}\|=1,
\boldsymbol{Ah}=\boldsymbol{0},~~~s.t.~ \|\boldsymbol{h}\|=1,
\end{equation}
which has a closed-form solution by Singular Value Decomposition (SVD). $\boldsymbol{A}$ is obtained by stacking all $\boldsymbol{a}_i$.

\noindent \textbf{GS Spatially-Varying Discrete Homography Field.}  In image stitching application, it is often safe to make zero-parallax assumption as long as the (non-planar) scene is far enough. However, it is also not uncommon that such assumption is violated to the extent that warping with just one global homography causes unpleasant misalignments. To address this issue, APAP~\cite{zaragoza2013projective} proposes to compute a spatially-varying homograpy field for each pixel $\boldsymbol{x}$:
\begin{equation}
\boldsymbol{h}^*(\boldsymbol{x}) = \arg\min_{\boldsymbol{h}} \sum_{i\in\mathcal{I}} \|w_i(\boldsymbol{x})\boldsymbol{a}_i\boldsymbol{h}\|^2,~~~s.t.~ \|\boldsymbol{h}\|=1,
\label{eq:gsapap}
\end{equation}
where $w_i(\boldsymbol{x}) = \max(\exp(-\frac{\|\boldsymbol{x}-\boldsymbol{x}_i\|^2}{\sigma^2}),\tau)$
is a weight. $\sigma$ and $\tau$ are the pre-defined scale and regularization parameters respectively. $\mathcal{I}$ indicates the inlier set returned from GS discrete 4-point solver with RANSAC (motivated by~\cite{tran2012defence}). The optimization has a closed-form solution by SVD. On the one hand, Eq.~\ref{eq:gsapap} encourages the warping to be globally As-Projective-As-Possible (APAP) by making use of all the inlier correspondences, while, on the other hand, it allows local deformations guided by nearby correspondences to compensate for model deficiency. Despite being a simple tweak, it yet leads to considerable improvement in image  stitching. %making APAP one of the most popular algorithms for stitching.  

\noindent \textbf{GS Differential Homography.}  %We now switch the focus to differential motion field as studied by Horn~\cite{horn1988motion}.
Suppose the camera is undergoing an instantaneous motion~\cite{horn1988motion}, consisting of rotational and translational velocity $(\boldsymbol{\omega},\boldsymbol{v})$. It would induce a motion flow $\boldsymbol{u}\in\mathbb{R}^2$ in each image point $\boldsymbol{x}$. Denoting $\tilde{\boldsymbol{u}}=[\boldsymbol{u}^\top, 0]^\top$, we have\footnote{See our supplementary material for derivations.}
\begin{equation}
\tilde{\boldsymbol{u}} = (\boldsymbol{I}-\hat{\boldsymbol{x}}\boldsymbol{e}_3^\top)\boldsymbol{H}\hat{\boldsymbol{x}},
\label{eq:motionfield}
\end{equation}
where $\boldsymbol{H}= -(\lfloor \boldsymbol{\omega} \rfloor_\times + {\boldsymbol{vn}^\top}/{d})$ is defined as the \emph{differential} homography~\cite{ma2012invitation}. $\boldsymbol{I}$ represents identity matrix and $\boldsymbol{e}_3 = [0,0,1]^\top$. $\lfloor.\rfloor_\times$ returns the corresponding skew-symmetric matrix from the vector. 
Each flow estimate $\{\boldsymbol{u}_i,\boldsymbol{x}_i\}$ gives two effective constraints out of the three equations included in Eq.~\ref{eq:motionfield}, denoted as $\boldsymbol{b}_i\boldsymbol{h}=\boldsymbol{u}_i$, where $\boldsymbol{b}_i\in \mathbb{R}^{2\times 9}$ can be computed from $\boldsymbol{x}_i$.
In \emph{GS differential 4-point solver}, with a minimal of 4 flow estimates, $\boldsymbol{H}$ can be computed by solving:
\begin{equation}
%\boldsymbol{h}^* = \arg\min_{\boldsymbol{h}} \sum_i \|\boldsymbol{b}_i\boldsymbol{h}-\boldsymbol{u}_i\|^2 = \arg\min_{\boldsymbol{h}} \|\boldsymbol{Bh}-\boldsymbol{U}\|^2, 
\boldsymbol{Bh}=\boldsymbol{U},
\end{equation}
which admits closed-form solution by pseudo inverse. $\boldsymbol{B}$ and $\boldsymbol{U}$ are obtained by stacking all $\boldsymbol{b}_i$ and $\boldsymbol{u}_i$, respectively.
Note that, we can only recover $\boldsymbol{H}_L = \boldsymbol{H}+\varepsilon \boldsymbol{I}$ with an unknown scale $\varepsilon$, because $\boldsymbol{B}$ has a one-dimensional null space. One can easily see this by replacing $\boldsymbol{H}$ in Eq.~\ref{eq:motionfield} with $\varepsilon\boldsymbol{I}$ and observing that the right hand side vanishes, regardless of the value of $\boldsymbol{x}$. $\varepsilon$ can be determined subsequently by utilizing the special structure of calibrated $\boldsymbol{H}$. However, this is not relevant in our paper since we focus on image stitching on general uncalibrated images.

\section{Methods}

\subsection{RS Motion Parameterization}

\begin{figure}[t]
	\centering
    \includegraphics[width=1.0\linewidth, trim = 55mm 85mm 110mm 0mm, clip]{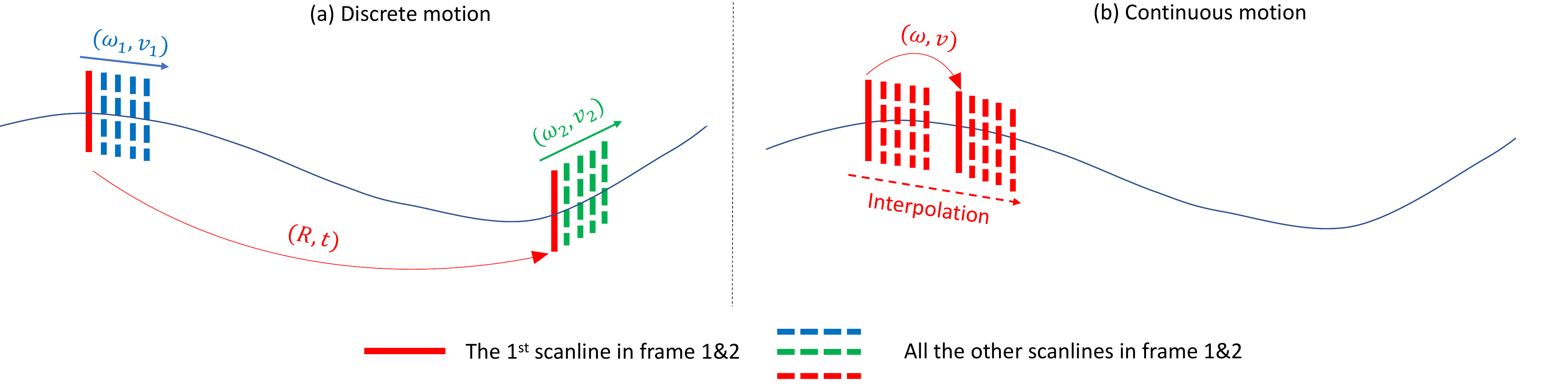}
	\caption{Illustration of discrete/continuous camera motion and their motion parameters.}
	\label{fig:rsmotion}
\end{figure}

Under the discrete motion model, in addition to the 6-Degree of Freedom (DoF) inter-frame relative motion $(\boldsymbol{R},\boldsymbol{t})$, 12 additional unknown parameters $(\boldsymbol{\omega}_1,\boldsymbol{v}_1)$ and $(\boldsymbol{\omega}_2,\boldsymbol{v}_2)$ are needed to model the intra-frame camera velocity, as illustrated in Fig.~\ref{fig:rsmotion}(a). This quickly increases the minimal number of points and the algorithm complexity to compute an RS-aware homography. Instead, we aim to solve for the case of continuous motion, i.e., a relatively small motion between two consecutive frames. In this case, we only need to parameterize the relative motion $(\boldsymbol{\omega},\boldsymbol{v})$ between the two first scanlines (one can choose other reference scanlines without loss of generality) of the image pair, and the poses corresponding to all the other scanlines can be obtained by interpolation, as illustrated in Fig.~\ref{fig:rsmotion}(b). In particular, it is shown in \cite{zhuang2017rolling} that a \emph{quadratic} interpolation can be derived under constant \emph{acceleration} motion. Formally, the absolute camera rotation and translation ($\boldsymbol{r}_1^{y_1}$,$\boldsymbol{p}_1^{y_1}$) (resp. ($\boldsymbol{r}_2^{y_2}$,$\boldsymbol{p}_2^{y_2}$)) of scanline $y_1$ (resp. $y_2$) in frame 1 (resp. 2) can be written as:
\begin{align}
    \boldsymbol{r}_1^{y_1} &= \beta_1(k,y_1)\boldsymbol{\omega},~~~\boldsymbol{p}_1^{y_1} = \beta_1(k,y_1)\boldsymbol{v},\\
    \boldsymbol{r}_2^{y_2} &= \beta_2(k,y_2)\boldsymbol{\omega},~~~\boldsymbol{p}_2^{y_2} = \beta_2(k,y_2)\boldsymbol{v},
\end{align}
where 
\begin{align}
\beta_1(k,y_1) & = (\frac{\gamma y_1}{h}+\frac{1}{2}k(\frac{\gamma y_1}{h})^2)(\frac{2}{2+k}),\\
\beta_2(k,y_2) & = (1+\frac{\gamma y_2}{h}+\frac{1}{2}k(1+\frac{\gamma y_2}{h})^2)(\frac{2}{2+k}).
\end{align}
Here, $k$ is an extra unknown motion parameter describing the acceleration, which is assumed to be in the same direction as velocity. $\gamma$ denotes the the readout time ratio~\cite{zhuang2017rolling}, i.e. the ratio between the time for scanline readout and the total time between two frames (including inter-frame delay).  $h$ denotes the total number of scanlines in a image. 
Note that the absolute poses ($\boldsymbol{r}_1^{y_1}$, $\boldsymbol{p}_1^{y_1}$) and ($\boldsymbol{r}_2^{y_2}$, $\boldsymbol{p}_2^{y_2}$) are all defined w.r.t the first scanline of frame 1. It follows that the relative pose between scanlines $y_1$ and $y_2$ reads:
\begin{align}
    \boldsymbol{\omega}_{y_1y_2} &= \boldsymbol{r}_2^{y_2} - \boldsymbol{r}_1^{y_1} = (\beta_2(k,y_2)-\beta_1(k,y_1))\boldsymbol{\omega},\\
    \boldsymbol{v}_{y_1y_2} &= \boldsymbol{p}_2^{y_2} - \boldsymbol{p}_1^{y_1} =  (\beta_2(k,y_2)-\beta_1(k,y_1))\boldsymbol{v}.
\end{align}
%If we fix $k=0$, it boils down to constant \emph{velocity} motion and the interpolation reduces to \emph{linear} interpolation. 
We refer the readers to \cite{zhuang2017rolling} for the detailed derivation of the above equations. 

\subsection{RS-Aware Differential Homography}
\label{sec:5pointsolver}
We are now in a position to derive the RS-aware differential homography.
First, it is easy to verify that Eq.~\ref{eq:motionfield} also applies uncalibrated cameras, under which case 
$\boldsymbol{H} = -\boldsymbol{K}(\lfloor \boldsymbol{\omega} \rfloor_\times + \boldsymbol{vn}^\top/d)\boldsymbol{K}^{-1}$, with $\boldsymbol{u}$ and $\boldsymbol{x}$ being raw measurements in pixels. $\boldsymbol{K}$ denotes the unknown camera intrinsic matrix.
Given a pair of correspondence by $\{\boldsymbol{u},\boldsymbol{x}\}$, we can plug $(\boldsymbol{\omega}_{y_1y_2},\boldsymbol{v}_{y_1y_2})$ into Eq.~\ref{eq:motionfield}, yielding %yielding the flow-homography relation as:
\begin{align}
\tilde{\boldsymbol{u}} &= (\beta_2(k,y_2) - \beta_1(k,y_1))(\boldsymbol{I}-\hat{\boldsymbol{x}}\boldsymbol{e}_3^\top)\boldsymbol{H}\hat{\boldsymbol{x}} = \beta(k,y_1,y_2)(\boldsymbol{I}-\hat{\boldsymbol{x}}\boldsymbol{e}_3^\top)\boldsymbol{H}\hat{\boldsymbol{x}}.
\label{eq:rsmotionfield}
\end{align}
Here, we can define $\boldsymbol{H}_{RS} = \beta(k,y_1,y_2)\boldsymbol{H}$ as the RS-aware differential homography, which is now scanline dependent.

\noindent \textbf{5-Point Solver.} In addition to $\boldsymbol{H}$, we now have one more unknown parameter $k$ to solve. Below, we show that 5 pairs of correspondences are enough to solve for $k$ and $\boldsymbol{H}$, using the so-called hidden variable technique~\cite{cox2006using}. To get started, let us first rewrite Eq.~\ref{eq:rsmotionfield} as:
\begin{equation}
\beta(k,y_1,y_2)\boldsymbol{b}\boldsymbol{h} = \boldsymbol{u}.
\end{equation}
Next, we move $\boldsymbol{u}$ to the left hand side and stack the constraints from 5 points, leading to:
\begin{equation}
\boldsymbol{C}\hat{\boldsymbol{h}} = \boldsymbol{0},
\end{equation}
where 
\begin{equation}
\boldsymbol{C} = \begin{bmatrix}
   \beta_1(k,y_1^1,y_2^1)\boldsymbol{b}_1, &~ -\boldsymbol{u}_1 \\
   \beta_2(k,y_1^2,y_2^2)\boldsymbol{b}_2, &~ -\boldsymbol{u}_2 \\
   \beta_3(k,y_1^3,y_2^3)\boldsymbol{b}_3, &~ -\boldsymbol{u}_3 \\
   \beta_4(k,y_1^4,y_2^4)\boldsymbol{b}_4, &~ -\boldsymbol{u}_4 \\
   \beta_5(k,y_1^5,y_2^5)\boldsymbol{b}_5, &~ -\boldsymbol{u}_5 \\
  \end{bmatrix}, 
  ~~~~~\hat{\boldsymbol{h}} = [\boldsymbol{h}^T, ~1]^T.
\end{equation}
It is now clear that, for $\boldsymbol{h}$ to have a solution, $\boldsymbol{C}$ must be rank-deficient. Further observing that $\boldsymbol{C} \in \mathbb{R}^{10\times 10}$ is a square matrix, rank deficiency indicates vanishing determinate, i.e.,  
\begin{equation}
    det(\boldsymbol{C})=\boldsymbol{0}.
\end{equation}
This gives a univariable polynomial equation, whereby we can solve for $k$ efficiently. $\boldsymbol{h}$ can subsequently be extracted from the null space of $\boldsymbol{C}$.

\noindent \textbf{DoF Analysis.} In fact, only 4.5 points are required in the minimal case, since we have one extra unknown $k$ while each point gives two constraints. Utilizing 5 points nevertheless leads to a straightforward solution as shown. \textit{Yet, does this lead to an over-constrained system}? No. Recall that we can only recover $\boldsymbol{H}+\varepsilon\boldsymbol{I}$ up to an arbitrary $\varepsilon$. Here, due to the one extra constraint, a specific value is chosen for $\varepsilon$ since the last element of $\hat{\boldsymbol{h}}$ is set to 1.
Note that a true $\varepsilon$, thus $\boldsymbol{H}$, is not required in our context since it does not affect the warping. % In other words, the ambiguity exists in 3D space but not in the image space. 
This is in analogy to uncalibrated SfM~\cite{hartley2003multiple} where a projective reconstruction up to an arbitrary projective transformation is not inferior to the Euclidean reconstruction in terms of reprojection error.

\noindent \textbf{Plane Parameters.} Strictly speaking, the plane parameters slightly vary as well due to the intra-frame motion. This is however not explicitly modeled in Eq.~\ref{eq:rsmotionfield}, due to two reasons. First, although the intra-frame motion is in a similar range as the inter-frame motion (Fig.~\ref{fig:rsmotion}(b)) and hence has a large impact in terms of motion, it induces merely a small perturbation to the absolute value of the scene parameters, which can be safely ignored (see supplementary for a more formal characterization). Second, we would like to keep the solver as simple as possible as along as good empirical results are obtained (see Sec.~\ref{sec:experiments}).

%\noindent \textbf{Remark on Normalization.}

\noindent \textbf{Motion Infidelity vs. Shutter Fidelity.}  Note that the differential motion model is always an approximation specially designed for small motion. This means that, unlike its discrete counterpart, its fidelity decreases with increasing motion. Yet, we are only interested in relatively large motion such that the RS distortion reaches the level of being visually unpleasant. Therefore, a natural and scientifically interesting question to ask is, whether the benefits from modeling RS distortion (Shutter Fidelity) are more than enough to compensate for the sacrifices due to the approximation in motion model (Motion Infidelity).
%---we may try very hard to model the RS effect using differential motion model, but only to find out that it underperforms the traditional GS discrete model in the end. 
Although a theoretical characterization on such comparison is out of the scope of this paper, via extensive experiments in Sec.~\ref{sec:experiments}, we fortunately observe that the differential RS model achieves overwhelming dominance in this competition.%, leading to superior performance in practice. 

\noindent \textbf{Degeneracy.} \textit{Are there different pairs of $k$ and $\boldsymbol{H}$ that lead to the same flow field $\boldsymbol{u}$}? Although such degeneracy does not affect stitching, it does make a difference to rectification (Sec.~\ref{sec:rsrectification}). We leave the detailed discussion to the supplementary, but would like the readers to be assured that such cases are very rare, in accordance with Horn~\cite{horn1988motion} that motion flow is hardly ambiguous.

\noindent \textbf{More Details.} Firstly, note that although $\{\boldsymbol{u},\boldsymbol{x}\}$ is typically collected from optical flow in classical works~\cite{ma2000linear,heeger1992subspace} prior to the advent of keypoint descriptors (e.g., \cite{lowe2004distinctive,rublee2011orb}), we choose the latter for image stitching for higher efficiency. Secondly, if we fix $k=0$, i.e., constant velocity model, $(\boldsymbol{\omega},\boldsymbol{v})$ could be solved using a linear 4-point minimal solver similar to the GS case. However, we empirically find its performance to be inferior to the constant acceleration model in shaking cameras, and shall not be further discussed here.
\subsection{RS-Aware Spatially-Varying Differential Homography Field}    
\label{sec:rshomographyfield}

%Here, we start to develop the spatially-varying RS differential homography field.

\noindent \textbf{Can GS APAP~\cite{zaragoza2013projective} Handle RS Distortion by Itself?} As aforementioned, the adaptive weight in APAP (Eq.~\ref{eq:gsapap}) permits local deformations to account for the local discrepancy from the global model. However, we argue that APAP alone is still not capable of handling RS distortion. The root cause lies in the GS homography being used --- although the warping of pixels near correspondences are less affected, due to the anchor points role of correspondences, the warping of other pixels still relies on the transformation propagated from the correspondences 
%(though nearby ones have larger impacts),
and thus the model being used does matter here. 
%Even in the scenes with rich textures where dense correspondences are accurate, the issue persists for those out-of-view pixels.

\noindent \textbf{RS-Aware APAP.} Obtaining a set of inlier correspondences $\mathcal{I}$ from our RS differential 5-point solver with RANSAC, we formulate the spatially-varying RS-aware homography field as:
\begin{equation}
\boldsymbol{h}^*(\boldsymbol{x}) = \arg\min_{\boldsymbol{h}} \sum_{i\in \mathcal{I}} \|w_i(\boldsymbol{x})(\beta(k,y_1,y_2)\boldsymbol{b}_i\boldsymbol{h} - \boldsymbol{u}_i)\|^2,
\label{eq:rsapap}
\end{equation}
where $w_i(\boldsymbol{x})$ is defined in Sec.~\ref{sec:background}. Since $k$ is a pure motion parameter independent of the scene, we keep it fixed in this stage for simplicity. Normalization strategy~\cite{hartley1997defense} is applied to $(\boldsymbol{u},\boldsymbol{x})$ for numerical stability.  We highlight that the optimization has a simple closed-form solution, yet is geometrically meaningful in the sense that it minimizes the error between the estimated and the observed flow $\boldsymbol{u}$. This stands in contrast with the discrete homography for which minimizing reprojection error requires nonlinear iterative optimization. In addition, we also observe higher stability from the differential model in cases of keypoints concentrating in a small region (see supplementary for discussions).
%For the sake of efficiency, instead of computing $\boldsymbol{H}$ for each pixel, we divide the image into a grid of cells and compute one local $\boldsymbol{H}$ for each cell. %Note that APAP remains straightforward in spite of the RS distortion modeling --- this again demonstrates the advantage of adopting the differential motion model. %In practice, we observe that APAP indeed often improves the stitching results. 
\subsection{RS Image Stitching and Rectification}
\label{sec:rsrectification}
Once we have the homography $\boldsymbol{H}$ (either a global one or a spatially-varying field) mapping from frame 1 to frame 2, we can warp between two images for stitching. %Unique here is that we could perform RS rectification to remove the distortion in both the individual images and the stitched image. 
Referring to Fig.~\ref{fig:rsmotion}(b) and Eq.~\ref{eq:rsmotionfield}, for each pixel $\boldsymbol{x}_1=[x_1, y_1]^\top$ in frame 1, we find its mapping $\boldsymbol{x}_2=[x_2,y_2]^\top$ in frame 2 by first solving for $y_2$ as:
\begin{align}
y_2 = y_1 + \lfloor (\beta_2(k,y_2)-\beta_1(k,y_1))(\boldsymbol{I}-\hat{\boldsymbol{x}}_1\boldsymbol{e}_3^\top)\boldsymbol{H}\hat{\boldsymbol{x}}_1 \rfloor_y,
\end{align}
which admits a closed-form solution. $\lfloor.\rfloor_y$ indicates taking the $y$ coordinate. $x_2$ can be then obtained easily with known $y_2$. 
Similarly, $\boldsymbol{x}_1$ could also be projected to the GS canvas defined by the pose corresponding to the first scanline of frame 1, yielding its rectified point $\boldsymbol{x}_{g1}$.
$\boldsymbol{x}_{g1}$ can be solved according to
\begin{equation} 
\boldsymbol{x}_1 = \boldsymbol{x}_{g1} +\lfloor\beta_1(k,y_1)(\boldsymbol{I}-\hat{\boldsymbol{x}}_{g1}\boldsymbol{e}_3^\top)\boldsymbol{H}\hat{\boldsymbol{x}}_{g1}\rfloor_{xy},
\end{equation}
where $\lfloor.\rfloor_{xy}$ indicates taking $x$ and $y$ coordinate.
%\noindent \textbf{Advantages over 3D Reconstruction-Based RS Rectification~\cite{zhuang2017rolling}.}
%One might notice that our approach for RS image rectification bears some resemblances to the method in \cite{zhuang2017rolling}, in the sense that both are using instantaneous motion models. However, 
%Here, we would like to highlight a few advantages of Homography-based approach over the 3D reconstruction-based approach in \cite{zhuang2017rolling}. 

%Note that, in comparison to the SfM-based method~\cite{zhuang2017rolling}, our rectification  requires just a set of sparse correspondences as opposed to dense optical flow, and hence is more amenable to high-efficiency processing in image/video editing.

\section{Experiments}
\label{sec:experiments}

\subsection{Synthetic Data}
\label{sec:syn_exp}
\noindent \textbf{Data Generation.}
First, we generate motion parameters $(\boldsymbol{\omega},\boldsymbol{v})$ and $k$ with desired constraints. For each scanline $y_1$ (resp. $y_2$) in frame 1 (resp. 2), we obtain its absolute pose as $(R(\beta_1(k,y_1)\boldsymbol{\omega}),\beta_1(k,y_1)\boldsymbol{v})$ (resp. $(R(\beta_2(k,y_2)\boldsymbol{\omega}),\beta_2(k,y_2)\boldsymbol{v})$). Here, $R(\boldsymbol{\theta}) = \exp(\boldsymbol{\lfloor\theta\rfloor_\times})$ with $\exp$: so(3) $\xrightarrow{}$ SO(3). %Note that the pose is defined in a discrete sense to respect the true geometry. % i.e. $\boldsymbol{X}_{y_1} = R(\beta_1\omega)(\boldsymbol{X}_1-\beta_1v)$ where $X_*$ indicates 3D point $\boldsymbol{X}$ in respective coordinate systems.
Due to the inherent depth-translation scale ambiguity, the magnitude of $\boldsymbol{v}$ is defined as the ratio between the translation magnitude and the average scene depth. The synthesized image plane is of size 720$\times1280$ with a $60^\circ$ horizontal Field Of View (FOV).
Next, we randomly generate a 3D plane, on which we sample 100 3D points within FOV. 
Finally, we project each 3D point $\boldsymbol{X}$ to the RS image. Since we do not know which scanline observes $\boldsymbol{X}$, we first solve for $y_1$ from the quadratic equation:
\begin{equation}
y_1 = \lfloor \pi(\boldsymbol{R}(\beta_1(k,y_1)\boldsymbol{\omega})(\boldsymbol{X}-\beta_1(k,y_1)\boldsymbol{v})) \rfloor_y,
\end{equation}
where $\pi([a,b,c]^\top){=}[a/c,b/c]^\top$. $x_1$ can then be obtained easily with known $y_1$. Likewise, we obtain the projection in frame 2.

\begin{figure}[t]
    \begin{minipage}[h]{0.73\textwidth}
        \centering
        \includegraphics[width=0.9\linewidth, trim = 20mm 14mm 20mm 10mm, clip]{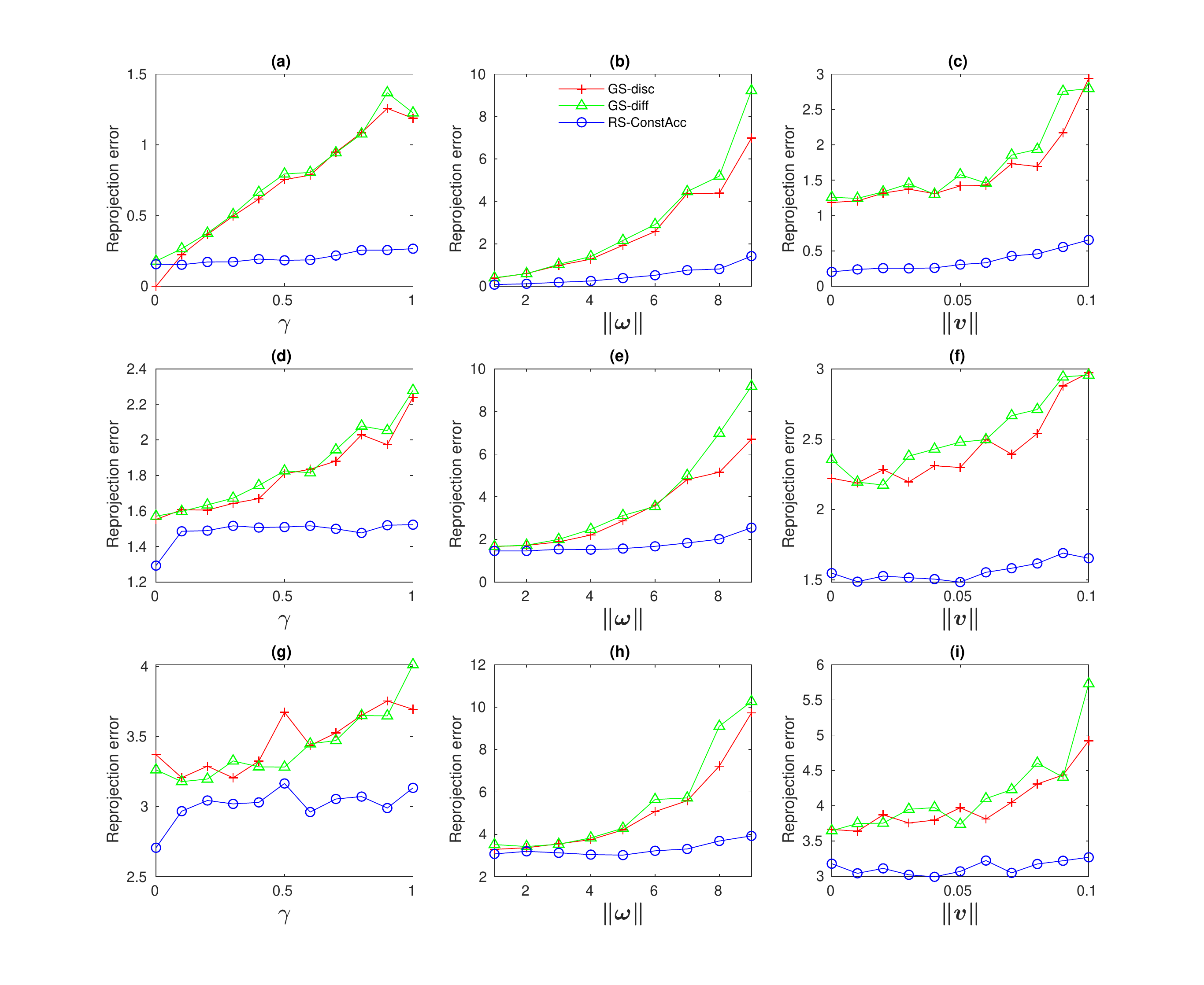}
        \caption{Quantitative comparison under different configurations.}
        \label{fig:syn1}
    \end{minipage}
    \begin{minipage}[h]{0.25\textwidth}
        \centering
        \includegraphics[width=0.95\linewidth, trim = 12mm 14mm 0mm 0mm, clip]{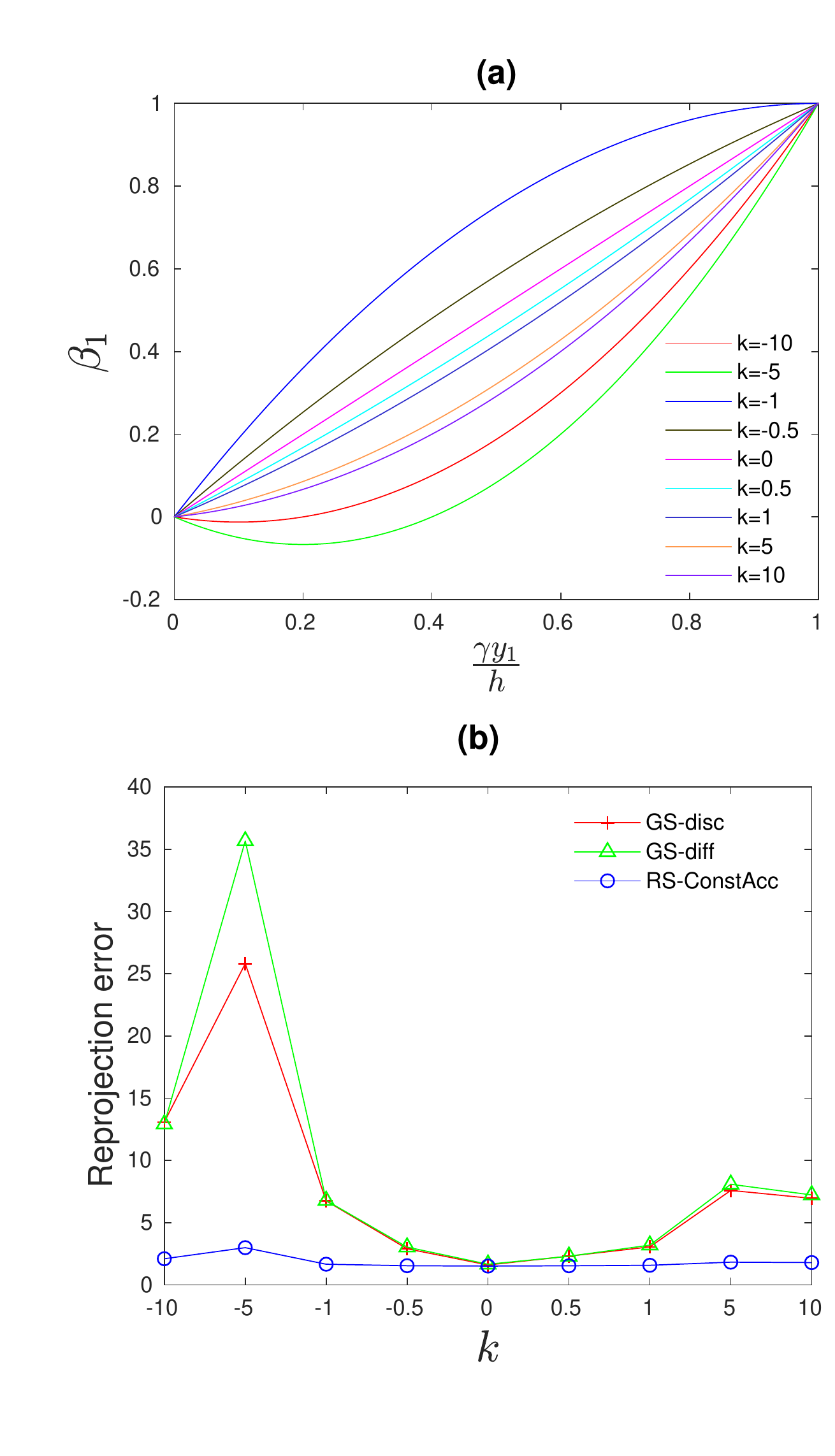}
        \caption{Quantitative comparison under different values of $k$.}
        \label{fig:syn2}
    \end{minipage}
\end{figure}

\noindent \textbf{Comparison under Various Configurations.} First, we study the performance under the noise-free case to understand the intrinsic and noise-independent behavior of different solvers, including discrete GS 4-point solver (`GS-disc'), differential GS 4-point solver (`GS-diff') and our RS 5-point solver (`RS-ConstAcc'). Specifically, we test the performance with varying RS readout time ratio $\gamma$, rotation magnitude $\|\boldsymbol{\omega}\|$, and translation magnitude $\|\boldsymbol{v}\|$. To get started, we first fix $(\|\boldsymbol{\omega}\|,\|\boldsymbol{v}\|)$ to $(3^\circ,0.03)$, and increase $\gamma$ from 0 to 1, indicating zero to strongest RS effect. Then, we fix $\gamma=1, \|\boldsymbol{v}\|=0.03$ while increasing $\|\boldsymbol{\omega}\|$ from $0^\circ$ to $9^\circ$. Finally, we fix $\gamma=1,\|\boldsymbol{\omega}\|=3^\circ$ while increasing $\|\boldsymbol{v}\|$ from 0 to 0.1. We report averaged reprojection errors over all point pairs in Fig.~\ref{fig:syn1}(a)-(c). The curves are averaged over 100 configurations with random plane and directions of $\boldsymbol{\omega}$ and $\boldsymbol{v}$.
%, each with 20 RANSAC trials to get best solution. 

First, we observe that `GS-diff' generally underperforms `GS-disc' as expected due to its approximate nature (cf. `Motion Infidelity' in Sec.~\ref{sec:5pointsolver}). In (a), although `RS-ConstAcc' performs slightly worse than `GS-disc' under small RS effect ($\gamma<=0.1$), it quickly surpasses `GS-disc' significantly with increasing $\gamma$ (cf. `Shutter Fidelity' in Sec.~\ref{sec:5pointsolver}). Moreover, this is constantly true in (b) and (c) with the gap becoming bigger with increasing motion magnitude. Such observations suggest that the gain due to handling RS effect overwhelms the degradation brought about by the less faithful differential motion model. Further, we conduct investigation with noisy data by adding Gaussian noise (with standard deviations $\sigma_g =1$ and $\sigma_g =2$ pixels) to the projected 2D points. The updated results in the above three settings are shown in Fig.~\ref{fig:syn1}(d)-(f) and Fig.~\ref{fig:syn1}(g)-(i) for $\sigma_g =1$ and $\sigma_g =2$ respectively. Again, we observe considerable superiority of the RS-aware model, demonstrating its robustness against noise. We also conduct evaluation under different values of $k$, with $(\|\boldsymbol{\omega}\|,\|\boldsymbol{v}\|)=(3^\circ,0.03)$, $\gamma=1$, $\sigma_g=1$. We plot $\beta_1(k,y_1)$ against $\frac{\gamma y_1}{h}$ with different values of $k$ in Fig.~\ref{fig:syn2}(a) to have a better understanding of scanline pose interpolation. The reprojection error curves are plotted in Fig.~\ref{fig:syn2}(b). We observe that the performance of `GS-disc' drops considerably with $k$ deviating from $0$, while `RS-ConstAcc' maintains almost constant accuracy. Also notice the curves are not symmetric as $k{>}0$ indicates acceleration (increasing velocity) while $k{<}0$ indicates deceleration (decreasing velocity).

\subsection{Real Data}
\label{sec:real_exp}
We find that the RS videos used in prior works, e.g.~\cite{grundmann2012calibration,ringaby2012efficient,hedborg2012rolling}, often contain small jitters without large viewpoint change across consecutive frames. To demonstrate the power of our method, we collect 5 videos (around 2k frames in total) with hand-held RS cameras while running, leading to large camera shaking and RS distortion. Following~\cite{Liu_2020_CVPR}, we simply set $\gamma=1$ to avoid its nontrival calibration~\cite{meingast2005geometric} and find it works well for our camera.

\begin{figure}[t]
	\centering
		\includegraphics[width=1.0\linewidth, trim = 0mm 380mm 43mm 0mm, clip]{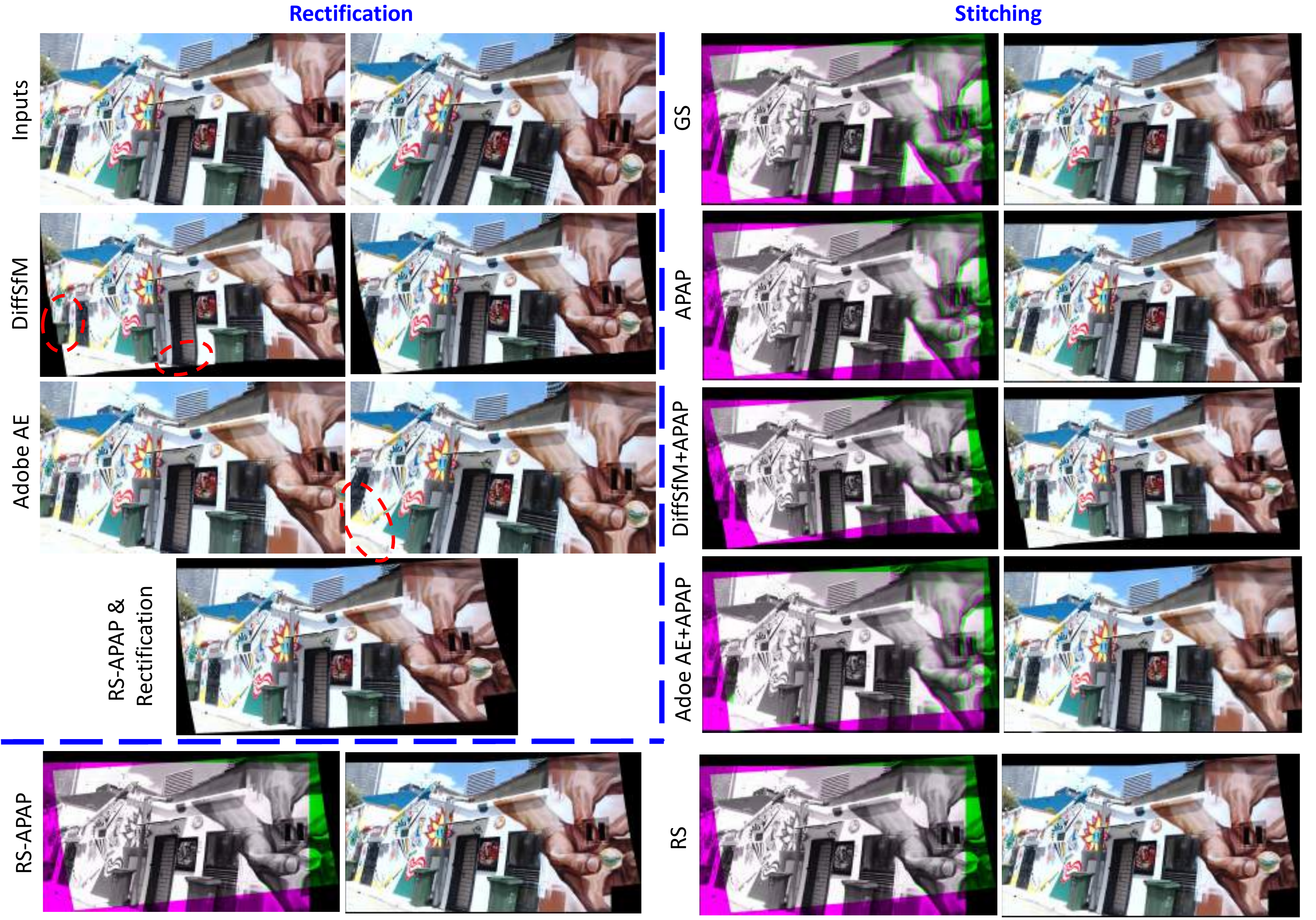}
	\centering
	\caption{Comparison of rectification/stitching on real RS images. Best viewed in screen.}
	\label{fig:twoviewexample1}
\end{figure}

\begin{figure}[t]
	\centering
		\includegraphics[width=1.0\linewidth, trim = 0mm 185mm 30mm 0mm, clip]{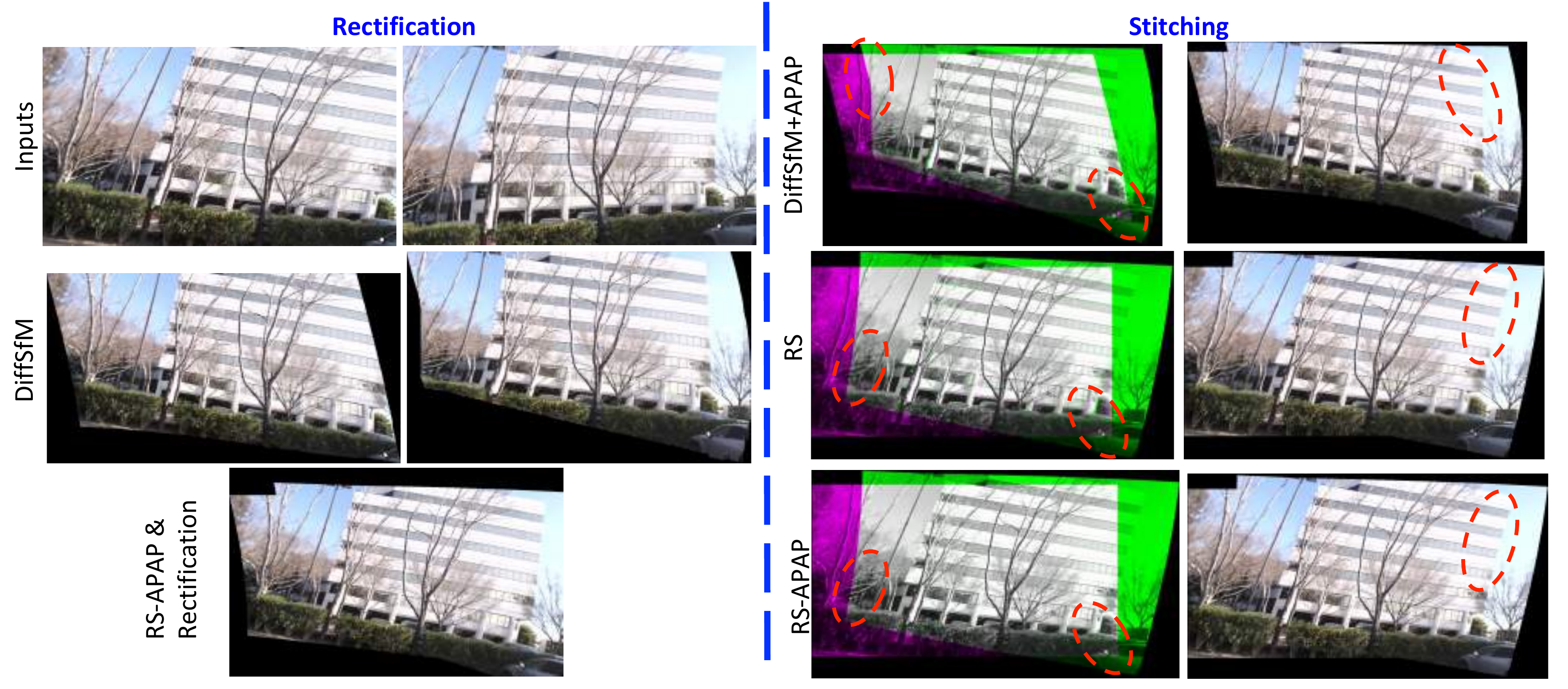}
	\centering
	\caption{Comparison of rectification/stitching on real RS images. Best viewed in screen.}
	\label{fig:twoviewexample2}
\end{figure}

\noindent \textbf{Two-View Experiments.} Below we discuss the two-view experiment results.

\noindent {\textit{Qualitative Evaluation.}}
We first present a few qualitative examples to intuitively demonstrate the performance gap, in terms of RS image rectification and stitching. For RS image rectification, we compare our method with the differential SfM based approach~\cite{zhuang2017rolling} (`DiffSfM') and the RS repair feature in Adobe After Effect (`Adobe AE'). For RS image stitching, we compare with the single GS discrete homography stitching (`GS') and its spatially-varying extension~\cite{zaragoza2013projective}(`APAP'). In addition, we also evaluate the sequential approaches which feed `DiffSfM' (resp. `Adobe AE') into `APAP', denoted as `DiffSfM+APAP' (resp. `Adobe AE+APAP'). We denote our single RS homography stitching without rectification as `RS', our spatially-varying RS homography stitching without rectification as `RS-APAP', and our spatially-varying RS homography stitching with rectification as `RS-APAP \& Rectification'.

In general, we observe that although `DiffSfM' performs very well for pixels with accurate optical flow estimates, it may cause artifacts elsewhere. Similarly, we find `Adobe AE' to be quite robust on videos with small jitters, but often introduces severe distortion with the presence of strong shaking. Due to space limit, we show two example results here and leave more to the supplementary. 

In the example of Fig.~\ref{fig:twoviewexample1}, despite that `DiffSfM' successfully rectifies the door and tube to be straight, the boundary parts (red circles) are highly skewed --- these regions have no correspondences in frame 2 to compute flow. `Adobe AE' manages to correct the images to some extent, yet bring evident distortion in the boundary too, as highlighted. `RS-APAP \& Rectification' nicely corrects the distortion with the two images readily stitched together. Regarding image stitching, we overlay two images after warping with the discrepancy visualized by green/red colors, beside which we show the linearly blended images. As can be seen, `GS' causes significant misalignments. `APAP' reduces them to some extent but not completely. The artifacts due to `DiffSfM' and `Adobe AE' persist in the stitching stage. Even for those non-boundary pixels, there are still misalignments as the rectification is done per frame in isolation, independent of the subsequent stitching. In contrast, we observe that even one single RS homography (`RS') suffices to warp the images accurately here, yielding similar result as `RS-APAP'.

We show one more example in Fig.~\ref{fig:twoviewexample2} with partial results (the rest are in the supplementary). `DiffSfM' removes most of the distortion to the extent that `APAP' warps majority of the scene accurately (`DiffSfM+APAP'), yet, misalignments are still visible as highlighted, again, due to its sequential nature. We would like to highlight that APAP plays a role here to remove the misalignment left by the `RS' and leads to the best stitching result.

\begin{figure}[t]
    \centering
    \begin{subfigure}
    \centering
		\includegraphics[width=0.38\linewidth, trim = 2mm 80mm 0mm 10mm, clip]{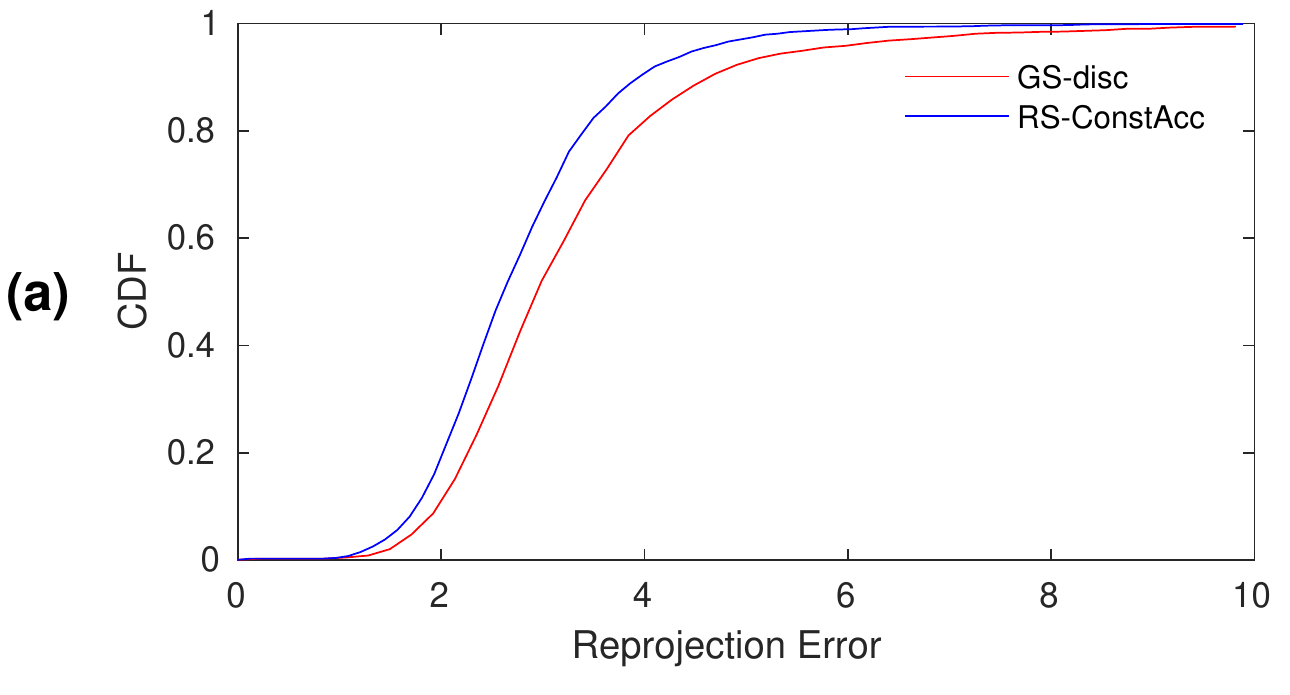}
	\end{subfigure}
	\begin{subfigure}
    \centering
		\includegraphics[width=0.38\linewidth, trim = 2mm 5mm 0mm 86mm, clip]{Figure/real1.pdf}
	\end{subfigure}
	\caption{Quantitative evaluation under standard setting \& further study.}
	\label{fig:real1}
\end{figure}

\noindent {\textit{Quantitative Evaluation.}}  Here, we conduct quantitative evaluation to characterize the benefits brought about by our RS model. For every pair of consecutive frames, we run both `GS-disc' and `RS-ConstAcc', each with 1000 RANSAC trials. We compute for each pair the median reprojection error among all the correspondences, and plot its  cumulative distribution function (CDF) across all the frame pairs, as shown in Fig.~\ref{fig:real1}(a). Clearly, `RS-ConstAcc' has higher-quality warping with reduced reprojection errors.

Although the above comparison demonstrates promising results in favor of the RS model, we would like to carry out further studies for more evidence, due to two reasons. First, note that the more complicated RS model has higher DoF and it might be the case that the smaller reprojection errors are simply due to over-fitting to the observed data, rather than due to truly higher fidelity of the underlying model. Second, different numbers (4 vs. 5) of correspondences are sampled in each RANSAC trial, leading to different amount of total samples used by the two algorithms. To address these concerns, we conduct two further investigations accordingly. First, for each image pair, we reserve 500 pairs of correspondences as test set and preclude them from being sampled during RANSAC. We then compare how well the estimated models perform on this set. Second, we test two different strategies to make the total number of samples equivalent --- `GS-MoreTrials': increases the number of RANSAC trials for `GS-disc' to $1000{\times}5/4=1250$; `GS-5point': samples non-minimal 5 points and get a solution in least squares sense in each trial. As shown in Fig.~\ref{fig:real1}(b), although `GS-5point' does improve the warping slightly, all the GS-based methods still lag behind the RS model, further validating the utility of our RS model.

\noindent \textbf{Comparison with Homographies for Video Stabilization~\cite{liu2013bundled,grundmann2012calibration}.} Here, we compare with the mesh-based spatially-variant homographies~\cite{liu2013bundled} and the homography mixture~\cite{grundmann2012calibration} proposed for video stabilization.
We would like to highlight that the fundamental limitation behind ~\cite{liu2013bundled,grundmann2012calibration} lies in that the individual homography is still GS-based, whereas ours explicitly models RS effect. We follow~\cite{li2015dual,lin2017direct} to evaluate image alignment by the RMSE of one minus normalized cross
correlation (NCC) over a neighborhood of $3{\times}3$ window for the overlapping pixel $\boldsymbol{x}_i$ and $\boldsymbol{x}_j$, i.e. $RMSE = \sqrt{\frac{1}{N}\sum_\pi(1-NCC(\boldsymbol{x}_i,\boldsymbol{x}_j))^2}$, with $N$ being the total number of pixels in the overlapping region $\pi$. As shown in Table.~\ref{tab:rmse}, RS-APAP achieves lower averaged RMSE than \cite{liu2013bundled,grundmann2012calibration}. Surprisingly, \cite{liu2013bundled} is not significantly better than GS, probably as its shape-preserving constraint becomes too strict for our strongly shaky videos. We also note that, in parallel with MDLT, our RS model could be integrated into~\cite{liu2013bundled,grundmann2012calibration} as well; this is however left as future works.

\begin{table}[t]
\caption{RMSE evaluation for image stitching using different methods.}
\begin{center}
\begin{tabular}{c||c|c|c|c|c}
  \hline
  Method      &~~~GS~~~&~~Mesh-based\cite{liu2013bundled}~~& ~~Mixture\cite{grundmann2012calibration}~~& ~~APAP\cite{zaragoza2013projective}~~&~~RS-APAP~~\\
  \hline
  RMSE([0-255]) &5.72	&5.15	&3.65&	3.27&	\textbf{3.05}
 \\
  \hline
\end{tabular}
\end{center}
\label{tab:rmse}
\end{table}

\begin{figure}[t]
    \centering
    \includegraphics[width=1.0\linewidth, trim = 0mm 80mm 140mm 0mm, clip]{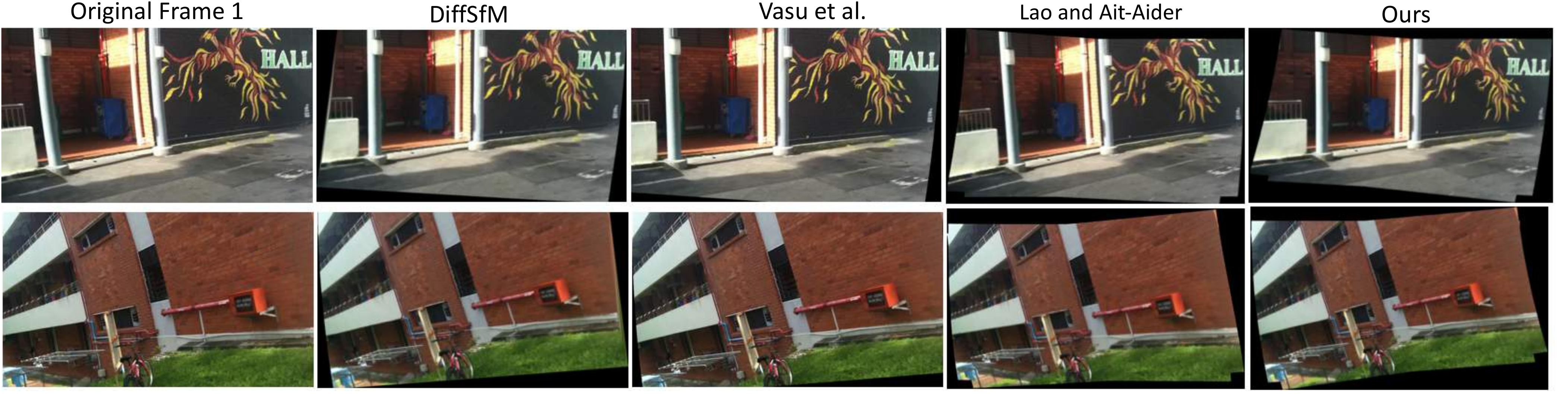}
	\caption{Qualitative comparison to DiffSfM~\cite{zhuang2017rolling}, the method of Lao and Ait-Aider~\cite{laopami20}, and the method of Vasu et al.~\cite{vasu2018occlusion}. Stitched images with rectification are shown for \cite{laopami20} and ours.}
	\label{fig:comparisontosfm}
\end{figure}
\noindent \textbf{Test on Data from \cite{zhuang2017rolling}.}
We also compare with ~\cite{laopami20,vasu2018occlusion} on the 6 image pairs used in \cite{zhuang2017rolling}, with 2 shown in Fig.~\ref{fig:comparisontosfm} and 4 in the supplementary. We show the results from our single RS model without APAP for a fair comparison to~\cite{zhuang2017rolling,laopami20}.
%First, although our approach is not based on full 3D reconstruction~\cite{zhuang2017rolling}, it can be seen that it is not worse in terms of visual appearance of the corrected images. 
First, we observe that our result is not worse than the full 3D reconstuction method~\cite{zhuang2017rolling}. 
In addition, it can be seen that our method performs on par with \cite{laopami20,vasu2018occlusion}, while being far more concise and simpler.

\begin{figure}[t]
    \centering
    \includegraphics[width=1.0\linewidth, trim = 0mm 50mm 0mm 0mm, clip]{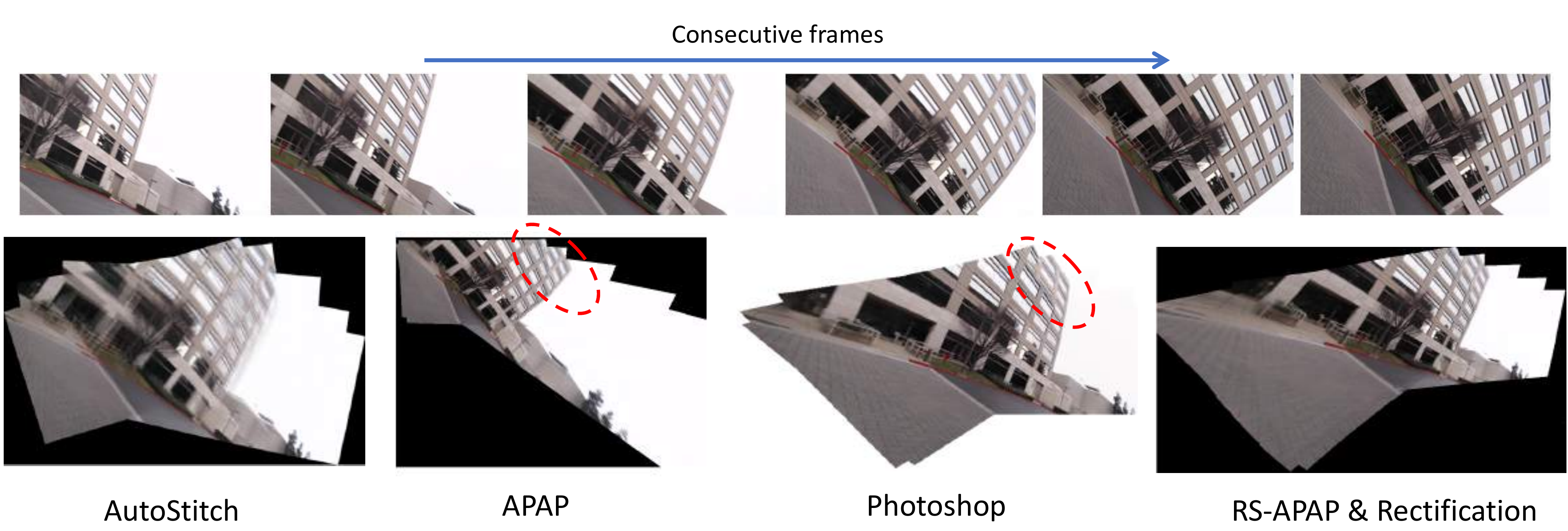}
	\caption{Qualitative comparison on multiple image stitching.}
    \label{fig:multi}
\end{figure}

\noindent \textbf{Multiple-View Experiments.}
%Although a more holistic way to handle multiple images jointly remains our future work, 
We demonstrate an extension to multiple images by concatenating the pairwise warping (note that the undertermined $\varepsilon$'s do not affect this step).
%In addition to the example in Fig.~\ref{fig:teaser}, 
We show an example in Fig.~\ref{fig:multi} and compare with the multi-image APAP~\cite{zaragoza2014projective}, AutoStitch~\cite{BL07} and Photoshop. AutoStitch result exhibits severe ghosting effects. APAP repairs them but not completely. Photoshop applies advanced seam cutting for blending, yet can not obscure the misalignments. Despite its naive nature, our simple concatenation shows superior stitching results.

\section{Conclusion}
\label{sec:conclusion}
We propose a new RS-aware differential homography together with its spatially-varying extension to allow local deformation. At its core is a novel minimal solver strongly governed by the underlying RS geometry. We demonstrate its application to  RS image stitching and rectification at one stroke, achieving good performance. 
%For future work, we would like to explore the more complex motion interpolation (in the line of \cite{ringaby2012efficient,grundmann2012calibration}) to handle more challenging motion, by either developing more minimal solvers or utilizing iterative optimization to further refine our current solution. Another direction for future works is to leverage our homography for 3D reconstruction and motion estimation, with applications in SfM/SLAM.
We hope this work could shed light on handling RS effect in other vision tasks such as large-scale SfM/SLAM~\cite{mur2015orb,schonberger2016structure,zhuang2018baseline,albl2016degeneracies,zhuang2019degeneracy}.

\noindent \textbf{Acknowledgements.} We would like to thank Buyu Liu, Gaurav Sharma, Samuel Schulter, and Manmohan Chandraker for proofreading and support of this work. We are also grateful to all the reviewers for their constructive suggestions.

\appendix
\section{Supplementary Material}
This supplementary material contains additional contents that are not discussed in the main paper due to space limit. Specifically, it includes: derivation of GS differential homography, derivation of RS differential homography that takes into account the varying plane parameters due to the intra-frame motion, degeneracy discussion, additional experimental results, and two demo videos for image rectification.
\subsection{Additional Details}
\noindent \textbf{Derivation of GS Differential Homography.} Here, we provide a brief derivation of motion flow in terms of GS differential homography as defined in Eq.~3 of the main paper. First, suppose the camera moves with the instantaneous velocity $(\boldsymbol{\omega},\boldsymbol{v})$, a 3D point $\boldsymbol{X} =[X,Y,Z]^\top$ in general position is observed by the camera to move with the velocity $\boldsymbol{V} = -\boldsymbol{v}-\lfloor\boldsymbol{\omega}\rfloor_\times\boldsymbol{X}$. Projecting this 3D velocity into the 2D image plane yields the motion flow $\boldsymbol{u}=[u_x,u_y]^\top$ at $(x,y)$~\cite{horn1988motion}
\begin{align}
u_x &= \frac{v_zx-v_x}{Z} -\omega_y +\omega_zy + \omega_xxy- \omega_yx^2,  \\
u_y &= \frac{v_zy-v_y}{Z} +\omega_x -\omega_zx - \omega_yxy + \omega_xy^2.
\label{eq:motionflow}
\end{align}
Since the 3D points $\boldsymbol{X}$ now lie in a plane, i.e. 
\begin{equation}
\boldsymbol{n}^\top\boldsymbol{X} =d, 
\end{equation}
and from perspective projection $\boldsymbol{X} = Z[x,y,1]^\top$ we have
\begin{equation}
    \frac{1}{Z} = \frac{\boldsymbol{n}^\top\hat{\boldsymbol{x}}}{d},
    \label{eq:Z}
\end{equation}
where $\hat{\boldsymbol{x}} = [x,y,1]^\top$.
Plugging Eq.~\ref{eq:Z} into Eq.~\ref{eq:motionflow}, it is trivial to verify that it becomes equivalent to the Eq.~3 in the main paper.

\noindent \textbf{Plane Parameters.} Here, we derive the differential homography that takes into account the varying plane parameters due to the intra-frame motion. For convenience, we will again adopt the instantaneous motion model. Specifically, the instantaneous camera velocity $(\boldsymbol{\omega},\boldsymbol{v})$ would induce an instantaneous rotation $\exp(-\lfloor\boldsymbol{\omega}\rfloor_\times)$ to the plane normal $\boldsymbol{n}$ and an instantaneous change $-\boldsymbol{v}^\top\boldsymbol{n}$ to the camera-plane distance $d$. When $(\boldsymbol{\omega},\boldsymbol{v})$ is applied to describe the motion between the two first scanlines of the image pair (Sec.~4.1 in the main paper), we can obtain the plane parameters w.r.t. the scanline $y_1$ in frame 1 as $(\exp(-\beta_1(k,y_1)\lfloor\boldsymbol{\omega}\rfloor_\times)\boldsymbol{n},d-\beta_1(k,y_1)\boldsymbol{v}^\top\boldsymbol{n})$. Combined with Eq.~9 and 10 in the main paper, we can write the homography between two scanlines $y_1$ and $y_2$ as 

\begin{equation}
    \boldsymbol{H}_{y_1y_2} = -\beta(k,y_1,y_2)\boldsymbol{K}\left(\lfloor\boldsymbol{\omega}\rfloor_\times +\frac{\boldsymbol{v}\left(\exp(-\beta_1(k,y_1)\lfloor\boldsymbol{\omega}\rfloor_\times)\boldsymbol{n}\right)^\top}{d-\beta_1(k,y_1)\boldsymbol{v}^\top\boldsymbol{n}}\right)\boldsymbol{K}^{-1}.
\end{equation}
For clarity, we shall omit the $k$, $y_1$ and $y_2$ for the derivation below. According to the first-order Taylor expansion, we have
\begin{align}
\exp(-\beta_1\lfloor\boldsymbol{\omega}\rfloor_\times) \approx \boldsymbol{I}-\beta_1\lfloor\boldsymbol{\omega}\rfloor_\times,\\
\frac{1}{d-\beta_1\boldsymbol{v}^\top\boldsymbol{n}} \approx \frac{1}{d}+\frac{\beta_1\boldsymbol{v}^\top\boldsymbol{n}}{d^2}.
\end{align}
Hence, we have 
\begin{align}
\frac{\boldsymbol{v}\left(\exp(-\beta_1\lfloor\boldsymbol{\omega}\rfloor_\times)\boldsymbol{n}\right)^\top}{d-\beta_1\boldsymbol{v}^\top\boldsymbol{n}} &= \frac{\boldsymbol{vn}^\top}{d}-\frac{\beta_1\boldsymbol{v}(\lfloor\boldsymbol{\omega}\rfloor_\times\boldsymbol{n})^\top}{d}+\\
& + \frac{\left(\beta_1\boldsymbol{v}^\top\boldsymbol{n}\right)\boldsymbol{vn}^\top}{d^2}- \frac{\left(\beta_1^2\boldsymbol{v}^\top\boldsymbol{n}\right)\boldsymbol{v}(\lfloor\boldsymbol{\omega}\rfloor_\times\boldsymbol{n})^\top}{d^2}.
\end{align}
Here, we can keep only the first-order term $\frac{\boldsymbol{vn}^\top}{d}$ and ignore other higher-order terms (in terms of $\boldsymbol{\omega}$ and $\boldsymbol{v}$) that have relatively small impact. We then have
\begin{equation}
    \boldsymbol{H}_{y_1y_2} = -\beta(k,y_1,y_2)\boldsymbol{K}\left(\lfloor\boldsymbol{\omega}\rfloor_\times+\frac{\boldsymbol{v}\boldsymbol{n}^\top}{d}\right)\boldsymbol{K}^{-1}.
\end{equation}
This is the differential RS-aware homography we use in the main paper; it allows us to derive the corresponding 5-point solver with only slightly higher complexity than its GS counterpart.

\noindent \textbf{Degeneracy.} Here, we would like to discuss in more details on whether there are different pairs of $k$ and $\boldsymbol{H}$ that lead to the same motion flow as defined in Eq.~11 of the main paper. Note that our interest is not in the well-known degeneracy~\cite{maybank2012theory,ma2012invitation} that $\boldsymbol{H}$ itself can arise from two valid pairs of motion and 3D plane, but rather in whether the RS effect induces additional degeneracy. Let us suppose $\{k_1,\boldsymbol{H}_1\}$ yields the same motion flow as does $\{k_2,\boldsymbol{H}_2\}$, i.e.,
\begin{equation}
\beta(k_1,y_1,y_2)(\boldsymbol{I}-\hat{\boldsymbol{x}}\boldsymbol{e}_3^\top)\boldsymbol{H}_1\hat{\boldsymbol{x}} = \beta(k_2,y_1,y_2)(\boldsymbol{I}-\hat{\boldsymbol{x}}\boldsymbol{e}_3^\top)\boldsymbol{H}_2\hat{\boldsymbol{x}}.
\label{eq:degeneracy}
\end{equation}
for all the $\boldsymbol{x}$ in the image plane.
Recall that $\beta(k,y_1,y_2) = (1+\frac{\gamma(y_2-y_1)}{h}+\frac{1}{2}k((1+\frac{\gamma y_2}{h})^2-(\frac{\gamma y_1}{h})^2))(\frac{2}{2+k})$. For a pair of $\boldsymbol{H}_1$ and $\boldsymbol{H}_2$, one can see that collecting the constraint Eq.~\ref{eq:degeneracy} from many points $\boldsymbol{x}$ quickly leads to an over-constrained system on $k_1$ and $k_2$. The equations cannot be all satisfied in general except under special configurations of motion and 3D plane. Here, we attempt to identify such configurations yet do not find any practical situations under which the degeneracy exists. Considering that $k_1$ and $k_2$ are just two independent scales, we reckon that such degeneracy, if existing at all, is very rare.
\subsection{Additional Results}
\begin{figure}[t]
	\begin{center}
		\includegraphics[width=1.0\linewidth, trim = 0mm 370mm 35mm 0mm, clip]{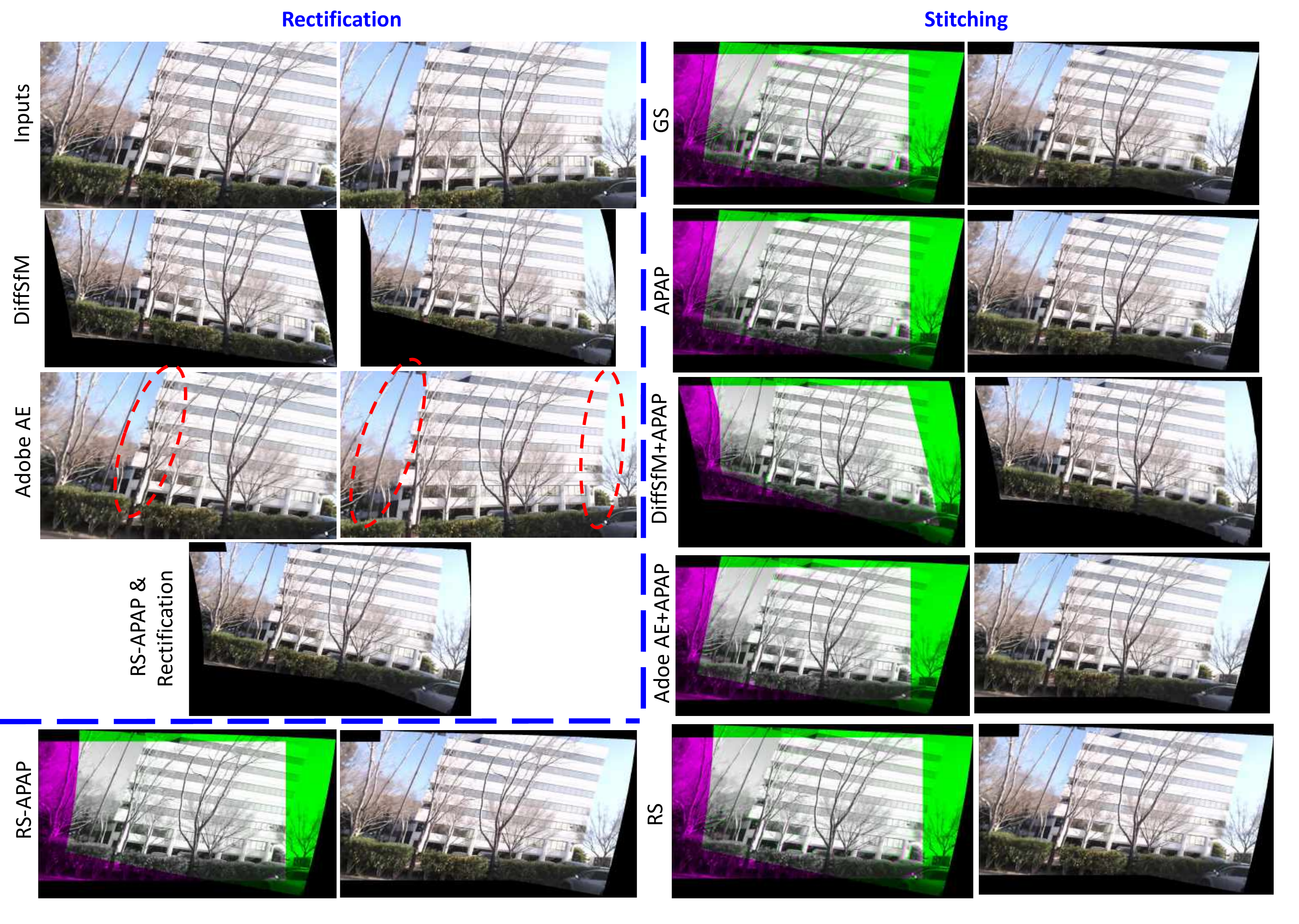}
	\end{center}
	%\vspace{-0.4cm}
	\caption{Comparison of rectification/stitching on real RS images. Best viewed in screen.}
	%\vspace{-0.4cm}
	\label{fig:twoviewexample0}
\end{figure}

\noindent \textbf{Two-View Experiments.} First, we present in Fig.~\ref{fig:twoviewexample0} the full results that we did not give in Fig. 6 of the main paper due to space constraints. We observe that `Adobe AE' has clearly inferior rectification results compared to both `DiffSfM' and `RS-APAP \& Rectification'. Second, we give additional two-view qualitative results in Fig.~\ref{fig:twoviewexample1} and \ref{fig:twoviewexample2}.
In the example shown in Fig.~\ref{fig:twoviewexample1}(a), we observe that both `DiffSfM' and `Adobe AE' introduces undesirable distortions in the image boundary. We reckon that the distortions in `Adobe AE' are perhaps due to its attempt to perform inpainting-like tasks in the boundary. However, it affects those pixels within the field of view as well. We have more examples later to demonstrate this observation. In the example shown in Fig.~\ref{fig:twoviewexample1}(b), we would like to highlight that, although all the three rectification methods are able to rectify the building to be mostly upright and it is hard to tell which performs best in this case, there are evident performance gap in the image stitching as highlighted. Especially, `DiffSfM' brings significant hindrance to the stitching (i.e., `DiffSfM+APAP' or \cite{zhuang2017rolling}+\cite{zaragoza2013projective}). This is presumably because of the discrepancy in the motion models --- the rectification applies the differential motion model while the stitching adopts the discrete one. Our `RS-APAP' obtains consistent results in rectification and stitching, thanks to its holistic nature. 
In the example shown in Fig.~\ref{fig:twoviewexample2}(a), we again observe that `DiffSfM' introduces undesirable distortions in the image boundary. The images returned by `Adobe AE' remain somewhat distorted; see the shape of the doors as highlighted. The superiority of `RS' and `RS-APAP' in image stitching is also clear. Example in Fig.~\ref{fig:twoviewexample2}(b) presents a challenging case, wherein the images contain large texture-less regions, i.e., windows, as well as slight motion blurs in frame 1. As expected, `DiffSfM' causes chaotic perturbation to these regions due to the difficulty in establishing correspondences. `Adobe AE' rectifies the images slightly, but again, produces large distortion in the boundary. `RS-APAP \& Rectification' rectifies the building to be upright, without introducing noticeable artifacts.

\noindent \textbf{Test on Data from~\cite{zhuang2017rolling}.} We give in Fig.~\ref{fig:comparisontosfm} the rectification results on the remaining four images from \cite{zhuang2017rolling}, which we did not provide in the main paper due to space limit. 
 
\begin{figure}
	\begin{center}
		\includegraphics[width=1.0\linewidth, trim = 0mm 110mm 35mm 0mm, clip]{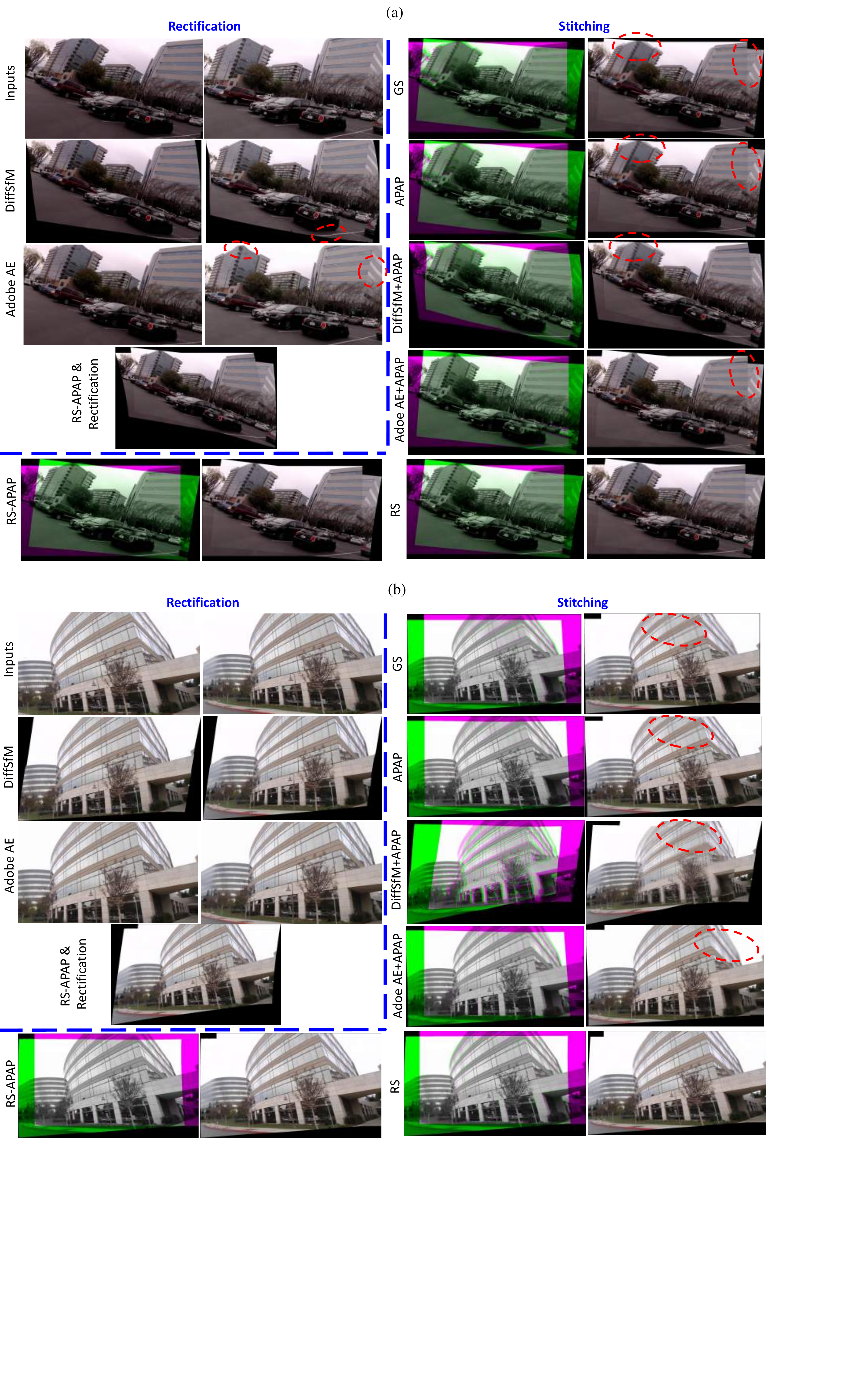}
	\end{center}
	\caption{Comparison of rectification/stitching on real RS images. Best viewed in screen.}
	\label{fig:twoviewexample1}
\end{figure}

\begin{figure}
	\begin{center}
		\includegraphics[width=1.0\linewidth, trim = 0mm 110mm 35mm 0mm, clip]{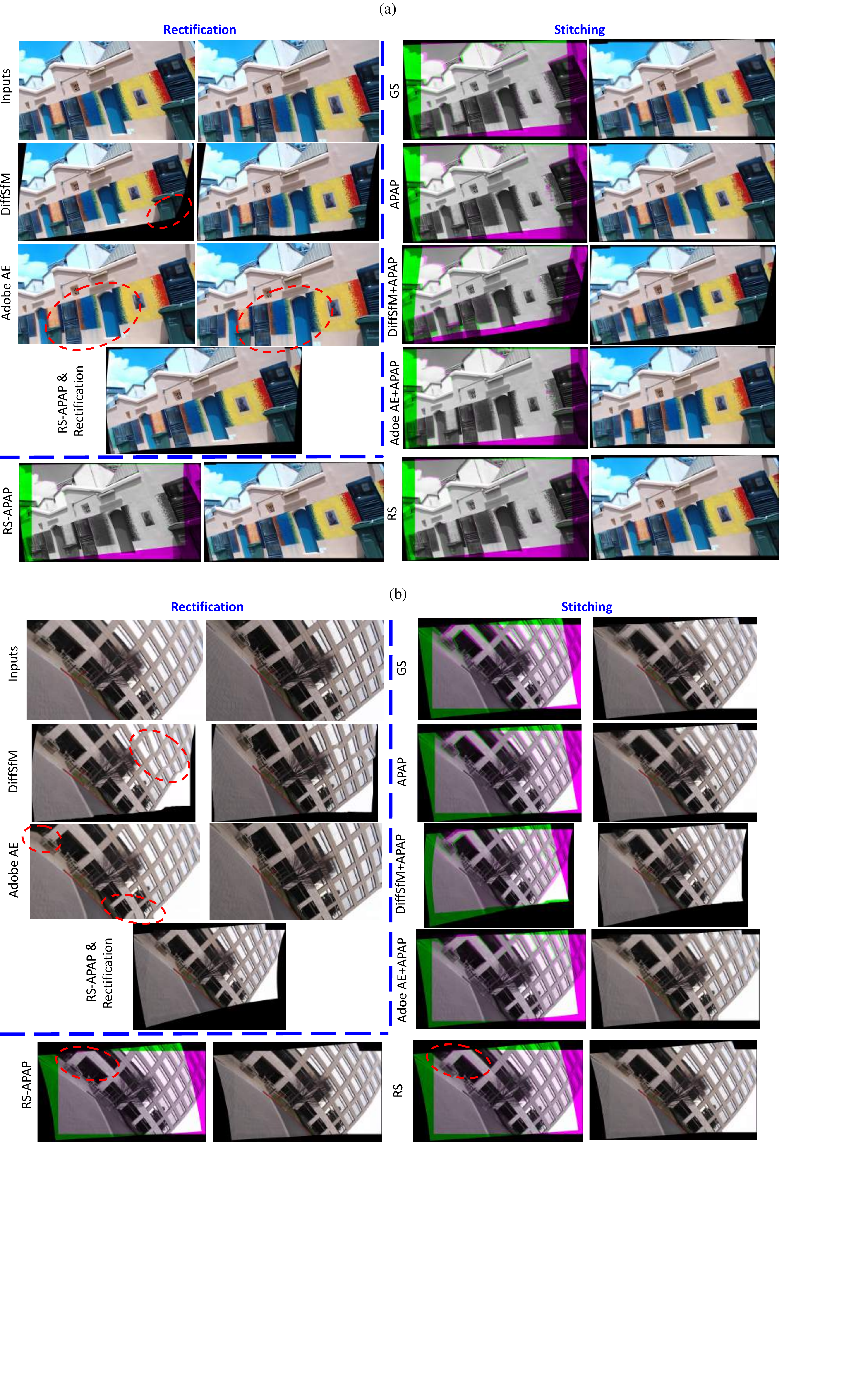}
	\end{center}
	%\vspace{-0.4cm}
	\caption{Comparison of rectification/stitching on real RS images. Best viewed in screen.}
	%\vspace{-0.4cm}
	\label{fig:twoviewexample2}
\end{figure}

\begin{figure}
	\begin{center}
		\includegraphics[width=1.0\linewidth, trim = 0mm 100mm 120mm 0mm, clip]{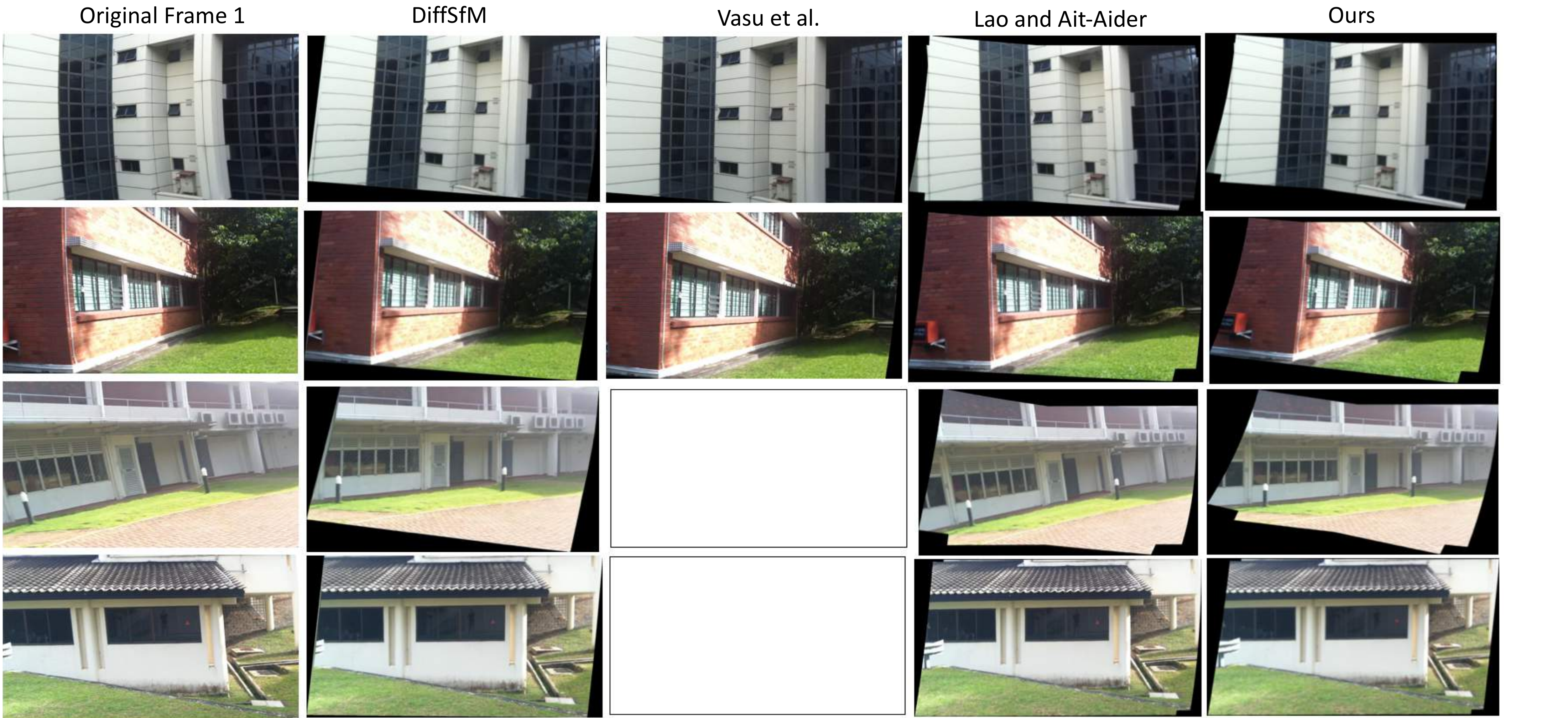}
	\end{center}
	%\vspace{-0.4cm}
	\caption{Qualitative comparison to DiffSfM~\cite{zhuang2017rolling}, the method of Lao and Ait-Aider~\cite{laopami20}, and the method of Vasu et al.~\cite{vasu2018occlusion}. Stitched images with rectification are shown for \cite{laopami20} and ours.~\cite{vasu2018occlusion}'s results for the last two images are not provided by the authors.}
	%\vspace{-0.4cm}
	\label{fig:comparisontosfm}
\end{figure}

\begin{figure}[t]
	\begin{center}
		\includegraphics[width=0.75\linewidth, trim = 0mm 0mm 140mm 0mm, clip]{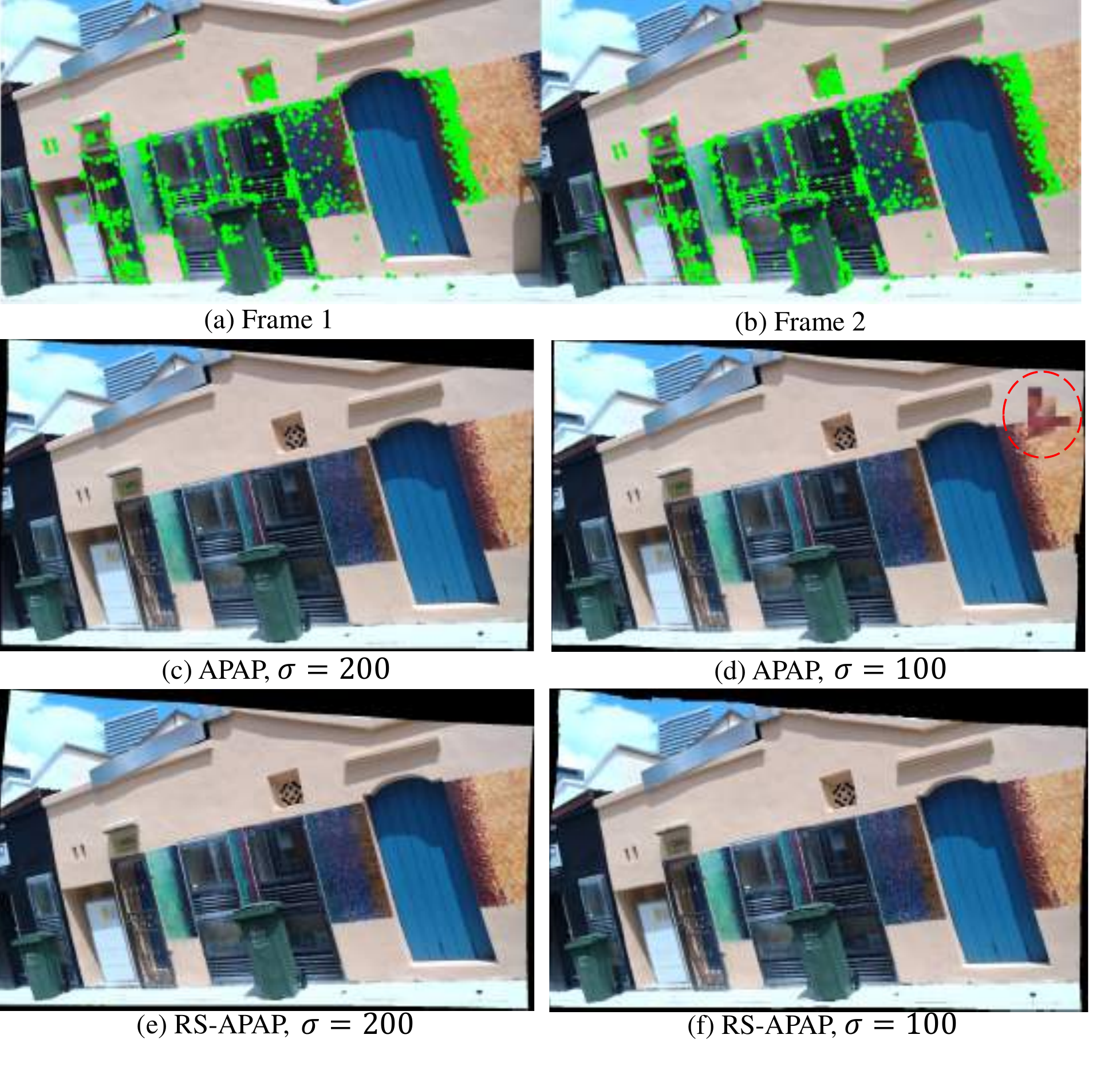}
	\end{center}
	%\vspace{-0.4cm}
	\caption{An example to demonstrate the sensitivity to keypoints distribution. (a)(b) show the original image pair with keypoints. (c)-(f) show the results of warping frame 2 to the canvas of frame 1 using APAP or RS-APAP with different values of $\sigma$.}
	%\vspace{-0.4cm}
	\label{fig:keypointdistribution}
\end{figure}

\noindent \textbf{Sensitivity to Keypoints Distribution.} As mentioned in the main paper, we observe that APAP might become unstable in cases of keypoints concentrating in a small region. Note that this is independent of the RS effect, and poses challenges to all keypoint-based methods. Nevertheless, we observe that our differential homography mitigates this issue to some extent as compared to the discrete homography. This may well be due to the higher stability of the differential motion model under small motion, which is exactly the motivation of its use in classical SfM~\cite{ma2000linear,ma2012invitation}. An example is shown in 
Fig.~\ref{fig:keypointdistribution}. As highlighted by the red circle, the APAP results might contain non-smooth artifacts near the small region where crowded keypoints are closely located in. This concentration of keypoints makes the local discrete homography estimation unstable, especially so with smaller value of $\sigma$. This further implies the hardness for APAP in such case to select a good value of $\sigma$ that can compensate for local deformation (smaller $\sigma$ desired) while leading to stable results. Our differential homography is more stable against the choice of $\sigma$.

\begin{figure}[t]
	\begin{center}
		\includegraphics[width=0.8\linewidth, trim = 0mm 5mm 150mm 0mm, clip]{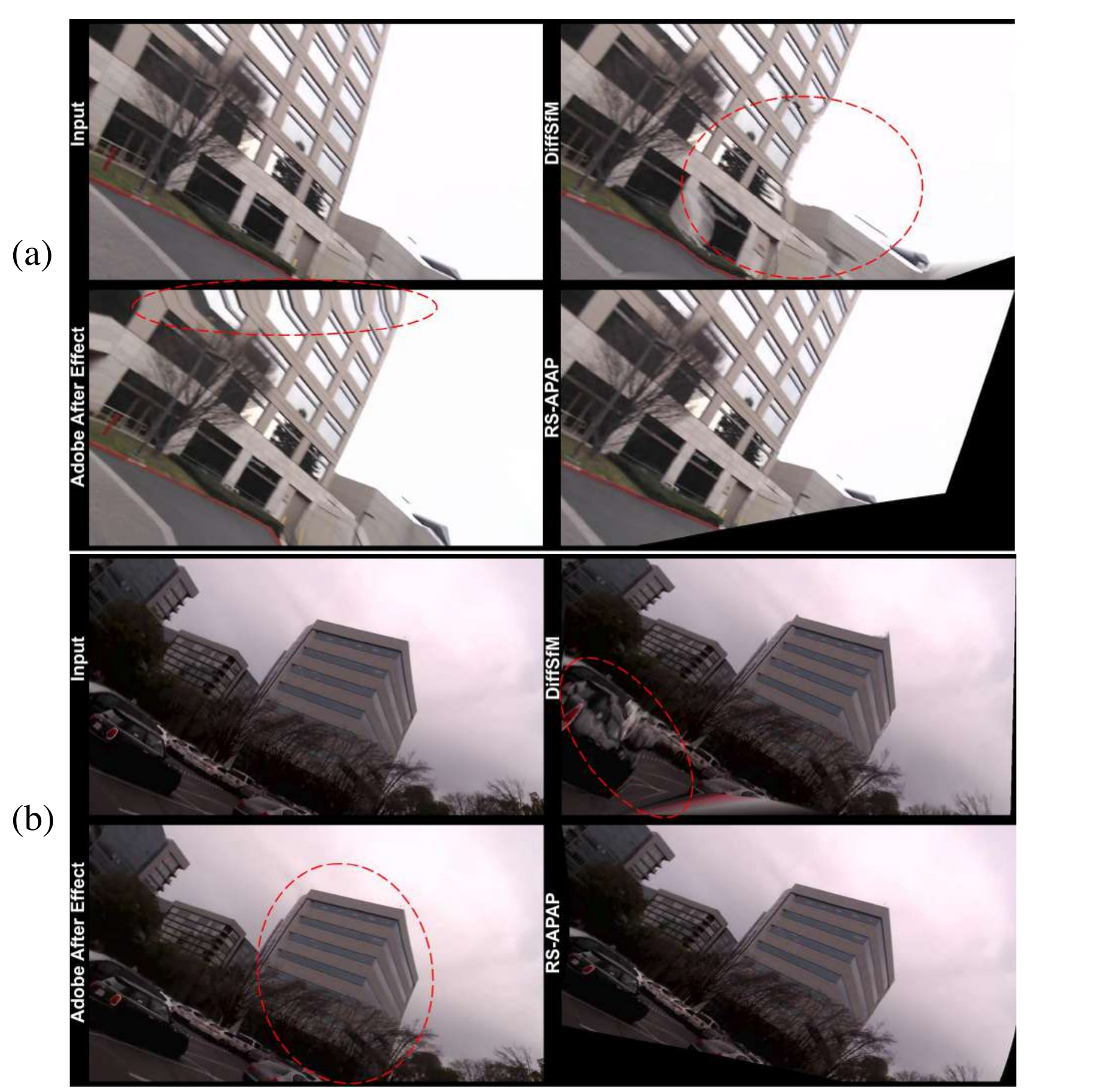}
	\end{center}
	\vspace{-0.4cm}
	\caption{Two examples extracted from the demo videos for image rectification.}
	\vspace{-0.4cm}
	\label{fig:demovideo}
\end{figure}

\begin{figure}[t]
	\begin{center}
		\includegraphics[width=0.85\linewidth, trim = 0mm 45mm 85mm 0mm, clip]{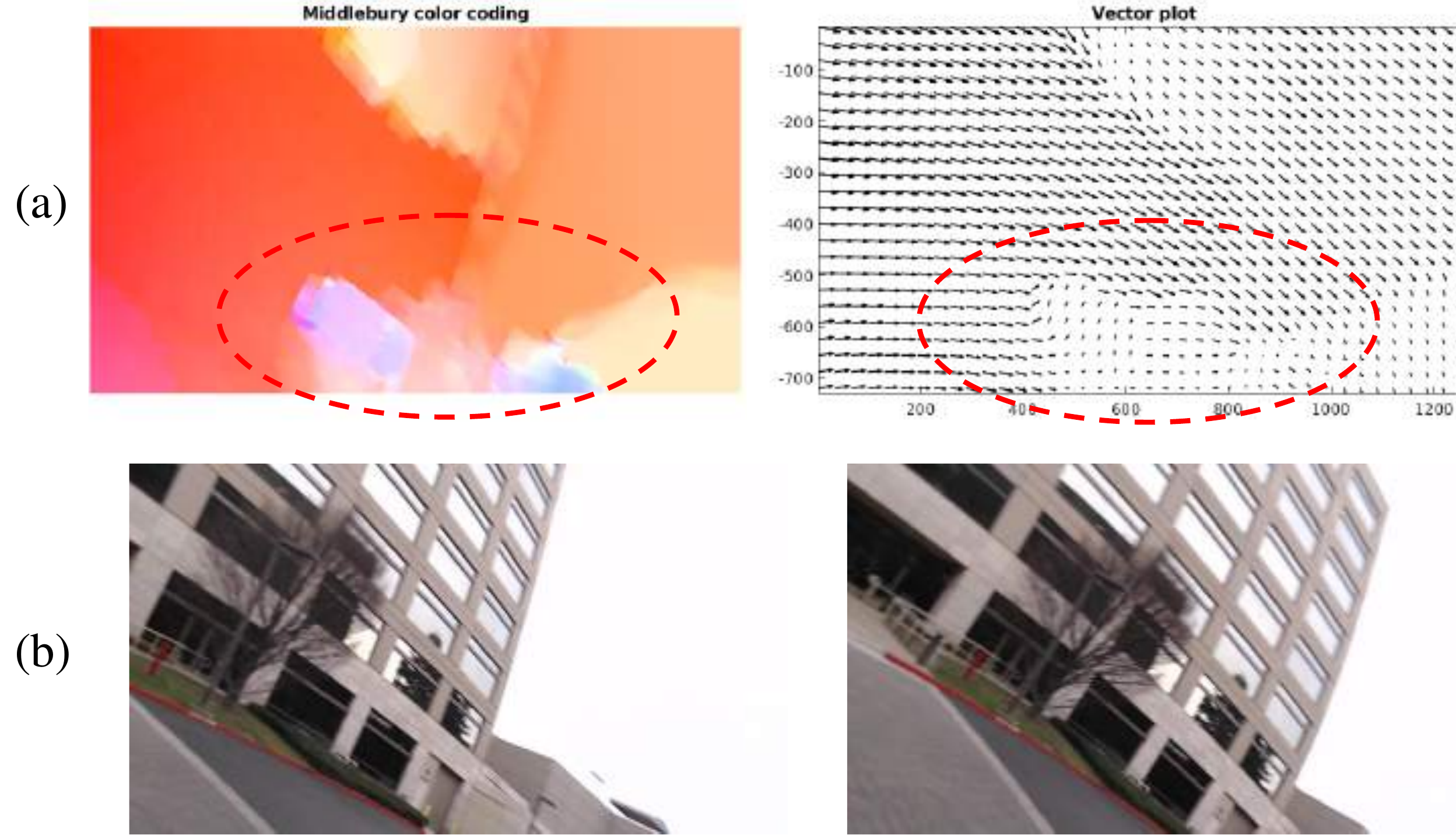}
	\end{center}
	\vspace{-0.4cm}
	\caption{(a) The optical flow results (visualized using \cite{sun2010secrets}) for the example in Fig.~\ref{fig:demovideo}(a). (b) The input image in Fig.~\ref{fig:demovideo}(a) and its next frame used to compute optical flow.}
	%\vspace{-0.4cm}
	\label{fig:flow}
\end{figure}

\noindent \textbf{Demo Videos.} We include two videos in this supplementary material to demonstrate the effectiveness of our image rectification method. The videos are captured by a hand-held RS camera while the holder is running, and hence contain large camera shaking, leading to significant image distortions. We compare our method with `DiffSfM' and `Adobe AE'. We refer the readers to the videos for the full results, and extract two frames from the videos as examples to show in Fig.~\ref{fig:demovideo}.
As can be seen in Fig.~\ref{fig:demovideo}(a), `DiffSfM' rectifies some parts of the scene while inducing evident distortions due to its sensitivity to errors in optical flow estimates. As visualized in Fig.~\ref{fig:flow},  it is obvious that the circled region contains unsmooth and wrong flow estimates, which inevitably affect the image rectification in `DiffSfM'. 
We can also see that `Adobe AE' strongly bends the straight lines near the boundary. In the example of Fig.~\ref{fig:demovideo}(b), `DiffSfM' again ruins some parts of the image, while `Adobe AE' fails to rectify the building as compared to our method. 
 
\clearpage
\bibliographystyle{splncs04}
\bibliography{reference}

\begin{thebibliography}{10}
\providecommand{\url}[1]{\texttt{#1}}
\providecommand{\urlprefix}{URL }
\providecommand{\doi}[1]{https://doi.org/#1}

\bibitem{albl2020two}
Albl, C., Kukelova, Z., Larsson, V., Polic, M., Pajdla, T., Schindler, K.: From
  two rolling shutters to one global shutter. In: CVPR (2020)

\bibitem{albl2015r6p}
Albl, C., Kukelova, Z., Pajdla, T.: R6p-rolling shutter absolute camera pose.
  In: CVPR (2015)

\bibitem{Albl_2016_CVPR}
Albl, C., Kukelova, Z., Pajdla, T.: Rolling shutter absolute pose problem with
  known vertical direction. In: CVPR (2016)

\bibitem{albl2016degeneracies}
Albl, C., Sugimoto, A., Pajdla, T.: Degeneracies in rolling shutter sfm. In:
  ECCV (2016)

\bibitem{bapat2018rolling}
Bapat, A., Price, T., Frahm, J.M.: Rolling shutter and radial distortion are
  features for high frame rate multi-camera tracking. In: CVPR (2018)

\bibitem{BL07}
Brown, M., Lowe, D.G.: Automatic panoramic image stitching using invariant
  features. International Journal of Computer Vision  \textbf{74}(1),  59--73
  (2007)

\bibitem{chang2014shape}
Chang, C.H., Sato, Y., Chuang, Y.Y.: Shape-preserving half-projective warps for
  image stitching. In: CVPR (2014)

\bibitem{chen2016natural}
Chen, Y.S., Chuang, Y.Y.: Natural image stitching with the global similarity
  prior. In: ECCV (2016)

\bibitem{cox2006using}
Cox, D.A., Little, J., O'shea, D.: Using algebraic geometry, vol.~185. Springer
  Science \& Business Media (2006)

\bibitem{dai2016rolling}
Dai, Y., Li, H., Kneip, L.: Rolling shutter camera relative pose: Generalized
  epipolar geometry. In: CVPR (2016)

\bibitem{grundmann2012calibration}
Grundmann, M., Kwatra, V., Castro, D., Essa, I.: Calibration-free rolling
  shutter removal. In: ICCP (2012)

\bibitem{haresh2020towards}
Haresh, S., Kumar, S., Zia, M.Z., Tran, Q.H.: Towards anomaly detection in
  dashcam videos. In: IV (2020)

\bibitem{hartley2003multiple}
Hartley, R., Zisserman, A.: Multiple view geometry in computer vision.
  Cambridge University Press (2003)

\bibitem{hartley1997defense}
Hartley, R.I.: In defense of the eight-point algorithm. IEEE Transactions on
  pattern analysis and machine intelligence  \textbf{19}(6),  580--593 (1997)

\bibitem{hedborg2012rolling}
Hedborg, J., Forss{\'e}n, P.E., Felsberg, M., Ringaby, E.: Rolling shutter
  bundle adjustment. In: CVPR (2012)

\bibitem{heeger1992subspace}
Heeger, D.J., Jepson, A.D.: Subspace methods for recovering rigid motion i:
  Algorithm and implementation. International Journal of Computer Vision
  \textbf{7}(2),  95--117 (1992)

\bibitem{herrmann2018robust}
Herrmann, C., Wang, C., Strong~Bowen, R., Keyder, E., Krainin, M., Liu, C.,
  Zabih, R.: Robust image stitching with multiple registrations. In: ECCV
  (2018)

\bibitem{herrmann2018object}
Herrmann, C., Wang, C., Strong~Bowen, R., Keyder, E., Zabih, R.:
  Object-centered image stitching. In: ECCV (2018)

\bibitem{horn1988motion}
Horn, B.K.: Motion fields are hardly ever ambiguous. International Journal of
  Computer Vision  \textbf{1}(3),  259--274 (1988)

\bibitem{im2015high}
Im, S., Ha, H., Choe, G., Jeon, H.G., Joo, K., So~Kweon, I.: High quality
  structure from small motion for rolling shutter cameras. In: ICCV (2015)

\bibitem{ito2017self}
Ito, E., Okatani, T.: Self-calibration-based approach to critical motion
  sequences of rolling-shutter structure from motion. In: CVPR (2017)

\bibitem{klingner2013street}
Klingner, B., Martin, D., Roseborough, J.: Street view
  motion-from-structure-from-motion. In: ICCV (2013)

\bibitem{kukelova2018linear}
Kukelova, Z., Albl, C., Sugimoto, A., Pajdla, T.: Linear solution to the
  minimal absolute pose rolling shutter problem. In: ACCV (2018)

\bibitem{laopami20}
Lao, Y., Aider, O.A.: Rolling shutter homography and its applications. In: IEEE
  Transactions on Pattern Analysis and Machine Intelligence (2020)

\bibitem{lao2018robust}
Lao, Y., Ait-Aider, O.: A robust method for strong rolling shutter effects
  correction using lines with automatic feature selection. In: CVPR (2018)

\bibitem{lao2018rolling}
Lao, Y., Ait-Aider, O., Bartoli, A.: Rolling shutter pose and ego-motion
  estimation using shape-from-template. In: ECCV (2018)

\bibitem{lee2020warping}
Lee, K.Y., Sim, J.Y.: Warping residual based image stitching for large
  parallax. In: CVPR (2020)

\bibitem{li2015dual}
Li, S., Yuan, L., Sun, J., Quan, L.: Dual-feature warping-based motion model
  estimation. In: ICCV (2015)

\bibitem{liao2019single}
Liao, T., Li, N.: Single-perspective warps in natural image stitching. IEEE
  Transactions on Image Processing  \textbf{29},  724--735 (2019)

\bibitem{lin2015adaptive}
Lin, C.C., Pankanti, S.U., Natesan~Ramamurthy, K., Aravkin, A.Y.: Adaptive
  as-natural-as-possible image stitching. In: CVPR (2015)

\bibitem{lin2016seagull}
Lin, K., Jiang, N., Cheong, L.F., Do, M., Lu, J.: Seagull: Seam-guided local
  alignment for parallax-tolerant image stitching. In: ECCV (2016)

\bibitem{lin2017direct}
Lin, K., Jiang, N., Liu, S., Cheong, L.F., Do, M., Lu, J.: Direct photometric
  alignment by mesh deformation. In: CVPR (2017)

\bibitem{lin2011smoothly}
Lin, W.Y., Liu, S., Matsushita, Y., Ng, T.T., Cheong, L.F.: Smoothly varying
  affine stitching. In: CVPR (2011)

\bibitem{liu2009content}
Liu, F., Gleicher, M., Jin, H., Agarwala, A.: Content-preserving warps for 3d
  video stabilization. ACM Transactions on Graphics (TOG)  \textbf{28}(3),
  ~1--9 (2009)

\bibitem{Liu_2020_CVPR}
Liu, P., Cui, Z., Larsson, V., Pollefeys, M.: Deep shutter unrolling network.
  In: CVPR (2020)

\bibitem{liu2013bundled}
Liu, S., Yuan, L., Tan, P., Sun, J.: Bundled camera paths for video
  stabilization. ACM Transactions on Graphics (TOG)  \textbf{32}(4),  1--10
  (2013)

\bibitem{lowe2004distinctive}
Lowe, D.G.: Distinctive image features from scale-invariant keypoints.
  International Journal of Computer Vision  \textbf{60}(2),  91--110 (2004)

\bibitem{ma2000linear}
Ma, Y., Ko{\v{s}}eck{\'a}, J., Sastry, S.: Linear differential algorithm for
  motion recovery: A geometric approach. International Journal of Computer
  Vision  \textbf{36}(1),  71--89 (2000)

\bibitem{ma2012invitation}
Ma, Y., Soatto, S., Kosecka, J., Sastry, S.S.: An invitation to 3-d vision:
  from images to geometric models, vol.~26. Springer Science \& Business Media
  (2012)

\bibitem{magerand2012global}
Magerand, L., Bartoli, A., Ait-Aider, O., Pizarro, D.: Global optimization of
  object pose and motion from a single rolling shutter image with automatic
  2d-3d matching. In: ECCV (2012)

\bibitem{maybank2012theory}
Maybank, S.: Theory of reconstruction from image motion, vol.~28. Springer
  Science \& Business Media (2012)

\bibitem{meingast2005geometric}
Meingast, M., Geyer, C., Sastry, S.: Geometric models of rolling-shutter
  cameras. In: Workshop on Omnidirectional Vision, Camera Networks and
  Non-Classical Cameras (2005)

\bibitem{mohan2017going}
Mohan, M.M., Rajagopalan, A., Seetharaman, G.: Going unconstrained with rolling
  shutter deblurring. In: ICCV (2017)

\bibitem{mur2015orb}
Mur-Artal, R., Montiel, J.M.M., Tardos, J.D.: Orb-slam: a versatile and
  accurate monocular slam system. IEEE transactions on robotics
  \textbf{31}(5),  1147--1163 (2015)

\bibitem{muratov20163dcapture}
Muratov, O., Slynko, Y., Chernov, V., Lyubimtseva, M., Shamsuarov, A., Bucha,
  V.: 3dcapture: 3d reconstruction for a smartphone. In: CVPRW (2016)

\bibitem{oth2013rolling}
Oth, L., Furgale, P., Kneip, L., Siegwart, R.: Rolling shutter camera
  calibration. In: CVPR (2013)

\bibitem{punnappurath2015rolling}
Punnappurath, A., Rengarajan, V., Rajagopalan, A.: Rolling shutter
  super-resolution. In: ICCV (2015)

\bibitem{purkait2018minimal}
Purkait, P., Zach, C.: Minimal solvers for monocular rolling shutter
  compensation under ackermann motion. In: WACV (2018)

\bibitem{purkait2017rolling}
Purkait, P., Zach, C., Leonardis, A.: Rolling shutter correction in manhattan
  world. In: ICCV (2017)

\bibitem{rengarajan2017unrolling}
Rengarajan, V., Balaji, Y., Rajagopalan, A.: Unrolling the shutter: Cnn to
  correct motion distortions. In: CVPR (2017)

\bibitem{rengarajan2016bows}
Rengarajan, V., Rajagopalan, A.N., Aravind, R.: From bows to arrows: Rolling
  shutter rectification of urban scenes. In: CVPR (2016)

\bibitem{rengarajan2016image}
Rengarajan, V., Rajagopalan, A.N., Aravind, R., Seetharaman, G.: Image
  registration and change detection under rolling shutter motion blur. IEEE
  Transactions on Pattern Analysis and Machine Intelligence  \textbf{39}(10),
  1959--1972 (2016)

\bibitem{ringaby2012efficient}
Ringaby, E., Forss{\'e}n, P.E.: Efficient video rectification and stabilisation
  for cell-phones. International Journal of Computer Vision  \textbf{96}(3),
  335--352 (2012)

\bibitem{rublee2011orb}
Rublee, E., Rabaud, V., Konolige, K., Bradski, G.: Orb: An efficient
  alternative to sift or surf. In: ICCV (2011)

\bibitem{saurer2013rolling}
Saurer, O., Koser, K., Bouguet, J.Y., Pollefeys, M.: Rolling shutter stereo.
  In: ICCV (2013)

\bibitem{saurer2016sparse}
Saurer, O., Pollefeys, M., Hee~Lee, G.: Sparse to dense 3d reconstruction from
  rolling shutter images. In: CVPR (2016)

\bibitem{saurer2015minimal}
Saurer, O., Pollefeys, M., Lee, G.H.: A minimal solution to the rolling shutter
  pose estimation problem. In: IROS (2015)

\bibitem{schonberger2016structure}
Schonberger, J.L., Frahm, J.M.: Structure-from-motion revisited. In: CVPR
  (2016)

\bibitem{Schubert_2018_ECCV}
Schubert, D., Demmel, N., Usenko, V., Stuckler, J., Cremers, D.: Direct sparse
  odometry with rolling shutter. In: ECCV (2018)

\bibitem{schubert2018direct}
Schubert, D., Demmel, N., Usenko, V., Stuckler, J., Cremers, D.: Direct sparse
  odometry with rolling shutter. In: ECCV (2018)

\bibitem{sun2010secrets}
Sun, D., Roth, S., Black, M.J.: Secrets of optical flow estimation and their
  principles. In: CVPR. IEEE (2010)

\bibitem{szeliski2007image}
Szeliski, R., et~al.: Image alignment and stitching: A tutorial. Foundations
  and Trends{\textregistered} in Computer Graphics and Vision  \textbf{2}(1),
  1--104 (2007)

\bibitem{tran2012defence}
Tran, Q.H., Chin, T.J., Carneiro, G., Brown, M.S., Suter, D.: In defence of
  ransac for outlier rejection in deformable registration. In: ECCV (2012)

\bibitem{vasu2018occlusion}
Vasu, S., Mohan, M.M., Rajagopalan, A.: Occlusion-aware rolling shutter
  rectification of 3d scenes. In: CVPR (2018)

\bibitem{vasu2017camera}
Vasu, S., Rajagopalan, A.N., Seetharaman, G.: Camera shutter-independent
  registration and rectification. IEEE Transactions on Image Processing
  \textbf{27}(4),  1901--1913 (2017)

\bibitem{zaragoza2013projective}
Zaragoza, J., Chin, T.J., Brown, M.S., Suter, D.: As-projective-as-possible
  image stitching with moving dlt. In: CVPR (2013)

\bibitem{zaragoza2014projective}
Zaragoza, J., Chin, T.J., Tran, Q.H., Brown, M.S., Suter, D.:
  As-projective-as-possible image stitching with moving dlt. IEEE Transactions
  on Pattern Analysis and Machine Intelligence  \textbf{36}(7),  1285--1298
  (2014)

\bibitem{zhang2014parallax}
Zhang, F., Liu, F.: Parallax-tolerant image stitching. In: CVPR (2014)

\bibitem{zhuang2017rolling}
Zhuang, B., Cheong, L.F., Hee~Lee, G.: Rolling-shutter-aware differential sfm
  and image rectification. In: ICCV (2017)

\bibitem{zhuang2018baseline}
Zhuang, B., Cheong, L.F., Hee~Lee, G.: Baseline desensitizing in translation
  averaging. In: CVPR (2018)

\bibitem{zhuang2019learning}
Zhuang, B., Tran, Q.H., Ji, P., Cheong, L.F., Chandraker, M.: Learning
  structure-and-motion-aware rolling shutter correction. In: CVPR (2019)

\bibitem{zhuang2019degeneracy}
Zhuang, B., Tran, Q.H., Lee, G.H., Cheong, L.F., Chandraker, M.: Degeneracy in
  self-calibration revisited and a deep learning solution for uncalibrated
  slam. In: IROS (2019)

\end{thebibliography}
\end{document}